\documentclass[preprint,12pt]{elsarticle}

\usepackage{amssymb}
\usepackage{amsmath}
\usepackage{amsmath,amsfonts}
\usepackage{algpseudocode}
\usepackage{algorithm}
\algnewcommand\algorithmicforeach{\textbf{for each}}
\algdef{S}[FOR]{ForEach}[1]{\algorithmicforeach\ #1\ \algorithmicdo}
\usepackage{array}
\usepackage[caption=false]{subfig}
\usepackage{textcomp}
\usepackage{stfloats}
\usepackage{url}
\usepackage{verbatim}
\usepackage{graphicx}
\usepackage{bm}
\usepackage{booktabs} 
\usepackage{multirow}
\usepackage{adjustbox}
\usepackage{threeparttable}
\usepackage{makecell}
\usepackage{color}
\usepackage{balance}
\usepackage[citecolor=green]{hyperref}
\usepackage{rotating}

\journal{Knowledge-Based Systems}

\begin{document}

\begin{frontmatter}



\title{Knowledge-aware Evolutionary Graph Neural Architecture Search}


\author{Chao Wang, Jiaxuan Zhao, Lingling Li$^*$, Licheng Jiao, Fang Liu, Xu Liu, and Shuyuan Yang} 

\affiliation{organization={School of Artificial Intelligence, Xidian University},
            city={Xian},
            postcode={710071}, 
            country={China}}
\begin{abstract}
Graph neural architecture search (GNAS) can customize high-performance graph neural network architectures for specific graph tasks or datasets. \textcolor{black}{However, existing GNAS methods begin searching for architectures from a zero-knowledge state, ignoring the prior knowledge that may improve the search efficiency.} The available knowledge base (e.g. NAS-Bench-Graph) contains many rich architectures and their multiple performance metrics, such as the accuracy (\#Acc) and number of parameters (\#Params). \textcolor{black}{This study proposes exploiting such prior knowledge to accelerate the multi-objective evolutionary search on a new graph dataset, named knowledge-aware evolutionary GNAS (KEGNAS).} KEGNAS employs the knowledge base to train a knowledge model and a deep multi-output Gaussian process (DMOGP) in one go, which generates and evaluates transfer architectures in only a few GPU seconds. The knowledge model first establishes a dataset-to-architecture mapping, which can quickly generate candidate transfer architectures for a new dataset. Subsequently, the DMOGP with architecture and dataset encodings is designed to predict multiple performance metrics for candidate transfer architectures on the new dataset. According to the predicted metrics, non-dominated candidate transfer architectures are selected to warm-start the multi-objective evolutionary algorithm for optimizing the \#Acc and \#Params on a new dataset. \textcolor{black}{Empirical studies on NAS-Bench-Graph and five real-world datasets show that KEGNAS swiftly generates top-performance architectures, achieving 4.27\% higher accuracy than advanced evolutionary baselines and 11.54\% higher accuracy than advanced differentiable baselines.} \textcolor{black}{In addition, ablation studies demonstrate that the use of prior knowledge significantly improves the search performance.}
\end{abstract}




\end{frontmatter}



\section{Introduction}
\label{sec1}
\textcolor{black}{Graph neural networks (GNN) are promising tools for learning representations of graph-structured data and are utilized in numerous real-world applications, such as robustness analysis, computer vision, and time-series prediction \cite{10026151}.} GNNs propagate feature information between neighboring nodes to learn deeper feature representations. Common GNNs include the graph convolutional network (GCN) \cite{kipf2017semisupervised}, graph sample and aggregate (GraphSAGE) \cite{hamilton2017inductive}, graph attention network (GAT) \cite{veličković2018graph}, and graph isomorphism network (GIN) \cite{xu2018how}. The main differences among these models lie in how they aggregate feature information between neighboring nodes and encode feature information \cite{9046288}. \textcolor{black}{Owing to their excellent performance and broad range of real-world applications, research on the general architectures of GNNs continues to grow.} The abundance of graph representation learning techniques makes it challenging to determine an ideal GNN model for a new graph task manually. This process is time consuming and laborious, necessitating significant expertise \cite{wang2021automated}.

Graph neural architecture search (GNAS) aims to automate the customization of the optimal GNN architecture for a given graph task, providing promising directions to address the above difficulties. GNAS typically involves exploring a large search space for possible GNN architectures, evaluating their performance on the task, and using optimization algorithms to determine the best architecture. This implies that the investigation of GNAS can be categorized into two central domains: search space and search strategy. Driven by these issues, researchers have developed numerous GNAS approaches to discover high-performance architectures that surpass the manually designed architectures for various graph tasks \cite{9782531,9714826,10.1145/3485447.3512185,GM2024112145,AN2023110341}. Unlike traditional convolutional neural networks (CNNs) that operate on structured data, the GNAS search space is defined on graphs of arbitrary sizes and shapes. Therefore, the search space of GNNs is significantly larger than that of CNNs. Moreover, because graphs typically contain numerous nodes and edges, scalable GNAS methods require further research \cite{wang2021automated}.

Despite its great success, GNAS still suffers from high computational costs and poor robustness. Existing work \cite{9458743,shala2023transfer} illustrates that one-shot or differentiable methods may be faster, but they lack robustness. Conversely, black-box optimization techniques such as evolutionary algorithms (EAs) \cite{9714826,SHI2022108752,10065594} and reinforcement learning (RL) \cite{10040227,9782531} are powerful, but require extensive evaluation. Transfer optimization has emerged as a promising area in the evolution community \cite{9321762}. Utilizing the experience or knowledge gained in previous optimization tasks to help the current optimization task provides a new perspective for achieving better optimization performance. Evolutionary transfer optimization (ETO) offers an increased optimizer’s capacity automatically with accumulated knowledge, motivating researchers to explore advanced optimizers with superior transferability \cite{9756606,9761797}. In recent years, a large-scale benchmark for GNAS (NAS-Bench-Graph) \cite{qin2022nasbenchgraph} has provided a wealth of architecture libraries and their multiple potentially conflicting performance metrics on benchmark datasets, such as the accuracy (\#Acc) and number of parameters (\#Params). Regrettably, the GNAS methods that are currently available cannot fully benefit from these valuable databases, which have the potential to improve the search efficiency significantly.

{In this paper, we propose the knowledge-aware multi-objective evolutionary GNAS (KEGNAS) framework to reduce the effort required to solve new graph tasks from scratch by efficiently capturing available knowledge from databases with multiple performance metrics, such as NAS-Bench-Graph. A knowledge model is trained once on a prior database consisting of benchmark datasets and architectures to learn a dataset-to-architecture mapping, which stores the available prior knowledge. For a new dataset, the knowledge model can generate numerous candidate transfer architectures. Subsequently, because evaluating the performance of a candidate transfer architecture on a new dataset is computationally expensive, a deep multi-output Gaussian process (DMOGP) with dataset and architecture encoding is constructed. The DMOGP predicts multiple performance metrics (\#Acc and \#Params) of the candidate transfer architectures on the new dataset. The surrogate model is trained once on a prior database consisting of benchmark datasets, architectures, and the performance of architectures on benchmark datasets. Notably, the generation and evaluation of candidate transfer architectures require only a few GPU seconds because of the avoidance of training on the new dataset. Finally, non-dominated candidate transfer architectures are selected as prior knowledge to warm-start the multi-objective EA (MOEA). The warm-start MOEA optimizes multiple performance metrics to obtain Pareto optimal architectures for the new graph dataset.

We perform the KEGNAS on the NAS-Bench-Graph and \textcolor{black}{five} real-world graph datasets. The experimental results reveal that KEGNAS outperforms several state-of-the-art GNAS methods. Moreover, ablation studies show the effectiveness of prior knowledge utilization under different task similarities. In summary, the key contributions of this study are as follows\footnote{Our code is publicly accessible at \url{https://github.com/xiaofangxd/KEGNAS}.}:

\begin{itemize}
  \item We provide an evolutionary transfer optimization perspective to improve the search efficiency of GNAS methods, which avoids the blindness of searching from the zero-knowledge state. The prior knowledge contained in NAS-Bench-Graph is leveraged to accelerate the search for high-performance architectures on new graph datasets.
  \item The novel KEGNAS framework is developed, which consists of three carefully designed components: a knowledge model, a DMOGP, and a warm-start MOEA. The knowledge model and DMOGP only require a few GPU seconds to generate and evaluate transfer architectures containing prior knowledge to assist multi-objective evolutionary searches on new datasets.
  \item The scalability and practicality of KEGNAS are demonstrated on NAS-Bench-Graph and real-world graph datasets. Ablation studies demonstrate that transfer architectures containing rich prior knowledge significantly improve the search performance.
\end{itemize}

The remainder of this paper is organized as follows. Section \ref{sec2} provides an overview of related studies on GNAS and ETO, emphasizing the motivating factors for our work. In addition, some techniques used in KEGNAS are presented. Section \ref{sec3} describes the problem formulation for KEGNAS. In Section \ref{sec4}, we describe the proposed KEGNAS framework in detail. In Section \ref{sec5}, the experimental results for the NAS-Bench-Graph and real-world graph datasets, along with ablation studies, are presented to demonstrate the scalability and utility of KEGNAS. Finally, we summarize our findings and potential directions for future research in Section \ref{sec6}.

\section{Preliminary}
\label{sec2}
In recent years, there has been a growing interest in GNAS. Because the technology of GNNs is widely employed in different computing scenarios, \textcolor{black}{searching the GNN architecture for a given graph task automatically is significant.} In this section, preliminary knowledge of GNNs is first introduced. We then provide a brief overview of GNAS methods closely related to the technical approach in this study according to two aspects: the search space and search strategy. Next, the related work on ETO is reviewed. Finally, we briefly introduce the MOGP, which is used in the proposed KEGNAS.

\subsection{Preliminary Knowledge about GNNs}

Let $\bm{G}=(\bm{V},\bm{E})$ be a graph-structured data, where $\bm{V}=\{v_1,...,v_{|V|}\}$ and $\bm{E}\subseteq \bm{V} \times \bm{V}$ represent the node and edge sets, respectively. $\mathcal{N}(i)=\{v_j\in \bm{V}|(v_i,v_j)\in \bm{E} \}$ is the neighborhood of node $v_i$. The node features are denoted as $\mathbf{X} = \{\mathbf{X}_i \in \bm{R}^{d}| v_i\in \bm{V}\}$.

GNNs are currently the most powerful tools for learning node and graph representations. As a state-of-the-art unified framework for GNNs, the message-passing network can be formalized as follows \cite{pmlr-v70-gilmer17a}:
\begin{equation}\label{eqGNN1}
    \mathbf{m}_i^{(k)}= \mathrm{AGG}^{(k)}\left(\left\{ \mathbf{W}^{(k)}\mathbf{h}_{j}^{(k-1)}, \forall j \in \mathcal{N}(i)\right\}\right),
\end{equation}
\begin{equation}\label{eqGNN2}
    \mathbf{h}_i^{(k)}=\mathrm{ACT}^{k}\left( \mathrm{f}^{(k)}\left(\mathbf{m}_i^{(k)}, \mathbf{h}_{i}^{(k-1)}\right) \right),
\end{equation}
where $\mathbf{m}_i^{(k)}$ and $\mathbf{h}_i^{(k)}$ are accordingly the message and node representations of node $v_i$ in the $k$th layer. $\mathrm{AGG}^{(k)}$, $\mathbf{W}^{(k)}$, $\mathrm{ACT}^{k}$, and $\mathrm{f}^{(k)}$ denote the aggregation operation, learnable weights, the activation function, and the combining function in layer $k$, respectively. In general, the node features $\mathbf{X}_i$ are employed to initialize the node representations $\mathbf{h}_i^{(0)}$. The final node representation after passing $K$ layers is defined as $\mathbf{H}^{(K)}=\{\mathbf{h}_i^{(K)}|v_i\in \bm{V}\}$. In addition, for graph-level tasks, the node representation $\mathbf{h}_{\bm{G}}$ of the entire graph $\bm{G}$ is aggregated using a readout operation $\mathrm{R}$:
\begin{equation}\label{eqGT}
    \mathbf{h}_{\bm{G}}=\mathrm{R}\left(\left\{ \mathbf{h}_{j}^{(K)}, \forall v_j \in \bm{V} \right\}\right).
\end{equation}

\subsection{Search Space}
The design of the search space determines the difficulty and scalability of GNAS. The existing search spaces for GNNs can mainly be divided into the following two categories: the micro search space and macro search space.

The micro search space is mainly employed to describe how to pass messages between the nodes in each layer. Taking the widely used messaging framework as an example, the common components include aggregation operations, combining functions, and activation functions \cite{qin2022nasbenchgraph}. Their main candidate operations are as follows:
\begin{itemize}
    \item Aggregation operations $\mathrm{AGG}$: GCN, GraphSAGE, GAT, GIN, ChebNet \cite{NIPS2016_04df4d43}, ARMA \cite{9336270}, and k-GNN \cite{Morris_Ritzert_Fey_Hamilton_Lenssen_Rattan_Grohe_2019},
    \item Combining function $\mathrm{f}$: MLP, CONCAT, and ADD.
    \item Activation function $\mathrm{ACT}$: Softplus, Sigmoid, Tanh, ReLU, Leaky ReLU, ReLU6, and ELU.
\end{itemize}

In addition, \textcolor{black}{the readout is considered in the micro search space to handle graph-level tasks.} Typical candidate operations include global pool, global gather, global attention pool, global attention sum pool, and flatten \cite{9378060}.

The macro-search space primarily describes the topology of GNNs \cite{10.1145/3485447.3512185}. Similar to the residual connections in deep learning, the node features of the current layer do not only depend on those of the previous layer \cite{Li_2019_ICCV}. Different connection designs also significantly influence the performance of GNNs. The connection schema is formalized as follows:
\begin{equation}\label{eqRC}
\mathbf{H}^{(k)}=\sum_{i<k} \mathrm{~S}_{i k}\left(\mathbf{H}^{(i)}\right),
\end{equation}
where $\mathrm{S}_{ik}$ indicates whether the node representation of the $i$th layer is selected as input at the $k$th layer. Common candidate operations include IDENTITY and ZERO, which represent ``select" and ``not select" respectively.

This study aims to use the prior knowledge in NAS-Bench-Graph to improve the performance of the GNAS method, so we further introduce the search space of NAS-Bench-Graph \cite{qin2022nasbenchgraph}. The search space can be expressed as a directed acyclic graph (DAG), in which the vertices and links are denoted by node representations and candidate operations, respectively. Each DAG contains an input vertex, an output vertex, and four intermediate vertices. Each intermediate vertex has only one incoming edge. Intermediate vertices with no successor vertices are connected to the output vertex via concatenation. The macro space of NAS-Bench-Graph contains eight DAGs, which we denote as $A, B, C, D, E, F, G, H, I$. In addition, nine candidate operations are considered: GCN, GraphSAGE, GAT, GIN, ChebNet, ARMA, k-GNN, residual connection (RC), and the fully connected layer (FC). After removing isomorphic structures, the entire search space includes 26206 architectures. Fig. \ref{fig1} presents an example of the visualization and encoding of a candidate architecture for NAS-Bench-Graph. A detailed description is provided in \cite{qin2022nasbenchgraph}.

\begin{figure}
\centering
\includegraphics[width=0.75\textwidth]{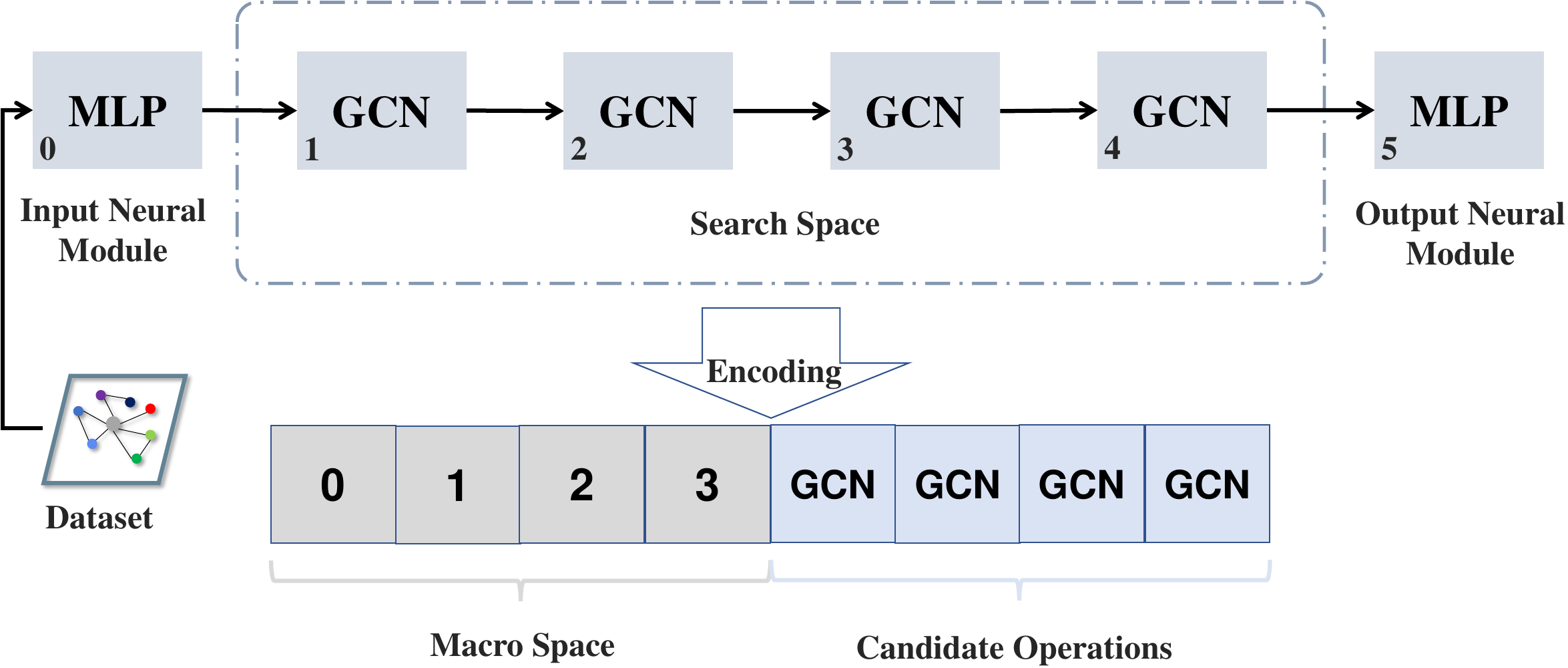}
\caption{An illustrative example of the GCN for NAS-Bench-Graph. The architecture encoding consists of the macro space and the candidate operations. In the macro space encoding, the value of the $i$th position represents the index of the previous vertex of vertex $i$. For example, the value of the 2nd position indicates that the index of the previous vertex of vertex 2 is 1. In the candidate operation encoding, the value of the $i$th position represents the candidate operation for vertex $i$. For example, the value of the 2nd position indicates that the candidate operation for vertex 2 is GCN.} \label{fig1}
\end{figure}

\subsection{Search Strategy}
The existing search strategies can be divided into three main categories: differentiable methods, RL-based methods, and EAs. Differentiable methods construct a super-network that contains all possible operations. Each operation is viewed as a probability distribution over all possible operations. The architecture and model parameters are alternately optimized using gradient-based algorithms \cite{9458743,liu2018darts}. For example, Zhao \textit{et al.} \cite{zhao2020probabilistic} first proposed a differentiable method for searching the macro-architecture and micro-architecture of GNNs. Wei \textit{et al.} \cite{10.1145/3485447.3512185} further explored the impact of the GNN topology on the model performance using a differentiable search method, which achieved state-of-the-art results on multiple real-world graph datasets. \textcolor{black}{Recently, Qin \textit{et al.} \cite{qin2023multitask} introduced a multi-task GNAS method that incorporates a layer-wise disentangled supernet capable of managing multiple architectures, significantly enhancing the performance by capturing task collaborations.}

RL-based methods train a controller via the RL policy to generate high-performance architectures \cite{10040227, zoph2017neural}. Zhou \textit{et al.} \cite{zhou2019auto} first designed an RL-based controller to validate architectures using small steps. This method uses a novel parameter-sharing strategy that enables homogeneous architectures to share parameters. Gao \textit{et al.} \cite{ijcai2020p195} trained recurrent neural networks using policy gradients to generate variable-length strings that describe neural network architectures. Gao \textit{et al.} \cite{9782531} extended the RL-based method to a distributed computing environment to improve the search efficiency. \textcolor{black}{Cai \textit{et al.} \cite{Cai_Wang_Li_Zhang_Zhu_2024} proposed an out-of-distribution multi-modal GNAS method, in which the controller and feature weights are iteratively optimized to ensure that the searched multi-modal model outputs representations devoid of spurious correlations.}

EAs continuously update a population to approximate the global optimum using genetic operators such as crossover, mutation, and selection. Because of their gradient-free and implicit parallelism, EAs have been employed to optimize neural network architectures for decades \cite{784219,9508774,10004638}. For GNAS, Shi \textit{et al.} \cite{SHI2022108752} first used a co-evolutionary algorithm (called Genetic-GNN) to search the architecture and hyperparameters of GNNs jointly to obtain an optimal model configuration. Nunes \textit{et al.} \cite{10.1145/3449639.3459318} employed fitness landscape analysis to measure the micro-search space of GNAS, showing that the fitness landscape is easy to explore. In addition, a parallel GNAS framework (Auto-GNAS) was proposed to improve the search efficiency of EAs through parallel computation \cite{9714826}. Liu et al. \cite{LIU2023110485} recently introduced a surrogate-assisted GNAS algorithm, known as CTFGNAS, to explore layer components, topology connections, and fusion strategies, while leveraging a surrogate model to reduce the computational cost. \textcolor{black}{Wang \textit{et al.} \cite{10681642} presented an automated configuration method for graph Transformer topologies and graph-aware strategies using a surrogate-assisted EA.}

Although GNAS is an intensively studied field, \textcolor{black}{no highly transferable method is yet available.} For a new graph task, existing methods can search only from scratch, which consumes considerable computing resources. Therefore, this study proposes the extraction of prior knowledge from previous databases to improve the efficiency of GNAS for an unseen new task.

\subsection{Evolutionary Transfer Optimization}
The existing methods for ETO can be divided into three categories\cite{xue2022}: evolutionary multitasking \cite{7161358,MA2023110027,GAO2024111530}, evolutionary multi-form optimization \cite{10026148}, and evolutionary sequential transfer optimization (ESTO) \cite{xue20221}. Reusing prior knowledge to solve new optimization tasks is known as sequential transfer optimization (STO), which has been applied in several fields, such as numerical optimization \cite{9385398,9756594} and multi-objective optimization (MOO) \cite{WANG2021107190,10.1162/evco_a_00300}. Next, we briefly review ESTO from two perspectives: model-based methods and solution-based methods \cite{xue2022}.

Model-based methods store and transfer knowledge by building a model. An intuitive idea is to use a prior knowledge base to construct a model to assist the current task \cite{9644585}; for example, the optimal probability distribution of the source task is used to initialize the search distribution of the current task \cite{Nomura_Watanabe_Akimoto_Ozaki_Onishi_2021}. Another concept involves aggregating the optimal probability distribution of the source task into the current task to construct a mixed distribution \cite{9761797,9950429}. Solution-based methods store and transfer knowledge in the form of solutions. For example, Feng \textit{et al.} \cite{7879282} proposed employing solutions to train denoising autoencoders to build the connections between tasks, where the solutions are regarded as carriers of knowledge. \textcolor{black}{Recently, Xue \textit{et al.} \cite{10342789} provided a comprehensive review of solution-based methods and explicitly explored the definition of solution transferability in the context of EAs. Scott \textit{et al.}. \cite{10.1145/3594805.3607137} presented important theoretical insights, including no-free-lunch theorems for transfer and the first asymptotic runtime analysis for transfer optimization.}

Transfer optimization has been used to address NAS in computer vision \cite{lee2021rapid,9328602,shala2023transfer}. Unlike images with a grid structure, graph-structured data lie in a non-Euclidean space, which leads to a unique design for graph learning \cite{wang2022automated}. Therefore, these methods are unsuitable for NAS problems with graph-structured data.

\subsection{Multi-output Gaussian Process}
Because multiple objectives can be modeled simultaneously, MOGP has been employed as a surrogate model for MOO problems in recent years \cite{pmlr-v97-astudillo19a,8836854}. A popular MOGP \cite{LIU2018102} can learn a multi-objective function $\bm{F}(\bm{x})=(f_1(\bm{x}),...,f_m(\bm{x}))^T$ by assuming the following:
\begin{equation}\label{MOGP_1}
\bm{y}=\begin{pmatrix}
 y_1\\
\vdots\\
 y_m\\
\end{pmatrix}=\begin{pmatrix}
 f_1(\bm{x})\\
\vdots\\
 f_m(\bm{x})\\
\end{pmatrix}+\begin{pmatrix}
 \epsilon_1\\
\vdots\\
 \epsilon_m\\
\end{pmatrix}=\bm{F}(\bm{x})+\bm{\epsilon},
\end{equation}
where $\epsilon_i\sim \mathcal{N}(0,\sigma^2_i), i=1,..,m$ is the Gaussian noise. The likelihood function can then be formulated as:
\begin{equation}\label{MOGP_2}
p(\bm{y}|\bm{F},\bm{x},\bm{\Sigma})=\mathcal{N}(\bm{F}(\bm{x}),\bm{\Sigma}),\bm{\Sigma}=\begin{pmatrix}
\sigma_1^2 & \cdots & 0 \\
\vdots & \ddots & \vdots \\
0 & \cdots & \sigma_m^2 \\
\end{pmatrix},
\end{equation}
where $\bm{\Sigma}\in \mathbb{R}^{m\times m}$ is a diagonal matrix. We assume that multiple objectives $\bm{F}(\bm{x})$ are drawn from a GP, as follows:
\begin{equation}\label{MOGP_3}
\begin{split}
&\bm{F}(\bm{x})\sim \mathcal{GP}(\bm{0},\mathbf{K}_w(\bm{x},\bm{x}^{'})),\\
& \mathbf{K}_w(\bm{x},\bm{x}^{'})=\begin{pmatrix}
k_{11}(\bm{x},\bm{x}^{'}) & \cdots & k_{1m}(\bm{x},\bm{x}^{'}) \\
\vdots & \ddots & \vdots \\
k_{m1}(\bm{x},\bm{x}^{'}) & \cdots & k_{mm}(\bm{x},\bm{x}^{'}) \\
\end{pmatrix},\\
\end{split}
\end{equation}
where $\mathbf{K}_w(\bm{x},\bm{x}^{'})\in\mathbb{R}^{m\times m}$ denotes the multi-output covariance. $k_{ij}(\bm{x},\bm{x}^{'})$ indicates the degree of similarity between the output $f_{i}(\bm{x})$ and $f_{j}(\bm{x}^{'})$. Given $N_{GP}$ inputs $\bm{X}=\{\bm{x}^{1},...,\bm{x}^{N_{gp}}\}$, the corresponding multiple objectives can be expressed as $\bm{Y}=(\bm{y}^{1},...,\bm{y}^{N_{gp}})^T=(y_1^{1},...,y_m^{1},...,y_1^{N_{gp}},...,y_m^{N_{gp}})^T$, where $y_i^{j}$ is the $i$th objective value on the $j$th input $\bm{x}^{j}$. For new data $\bm{x}^{'}$, the mean and covariance of the approximated $\bm{F}(\bm{x}^{'})$ are estimated as \cite{NEURIPS2021_a0d3973a}:
\begin{equation}\label{MOGP_4}
\begin{aligned}
& \bm{\mu}^{'}=\mathbf{K}_w(\bm{X},\bm{x}^{'})^T\left(\mathbf{K}_w(\bm{X},\bm{X})+\bm{\Sigma}^{'}\right)^{-1}\bm{Y} \\
& \begin{aligned}\bm{\Sigma}^{'}=&\mathbf{K}_w(\bm{x}^{'},\bm{x}^{'})-\\ &\mathbf{K}_w(\bm{X},\bm{x}^{'})^T\left(\mathbf{K}_w(\bm{X},\bm{X})+\bm{\Sigma}^{'}\right)^{-1}\mathbf{K}_w(\bm{X},\bm{x}^{'})\\
\end{aligned}\\
&
\end{aligned},
\end{equation}
where $\mathbf{K}_w(\bm{X},\bm{x}^{'})\in\mathbb{R}^{mN_{gp}\times m}$ has blocks
\begin{equation}\label{MOGP_5}
k_{ii^{'}}(\bm{X},\bm{x}^{'})=\begin{pmatrix}
k_{ii^{'}}(\bm{x}^1,\bm{x}^{'})\\
\vdots\\
 k_{ii^{'}}(\bm{x}^{N_{gp}},\bm{x}^{'})\\
\end{pmatrix},i,i^{'}=1,...,m.
\end{equation}

$\mathbf{K}_w(\bm{X},\bm{X})\in\mathbb{R}^{mN_{gp}\times mN_{gp}}$ has blocks
\begin{equation}\label{MOGP_6}
\begin{aligned}
& k_{ii^{'}}(\bm{X},\bm{X})=\begin{pmatrix}
k_{ii^{'}}(\bm{x}^1,\bm{x}^1) & \cdots & k_{ii^{'}}(\bm{x}^1,\bm{x}^{N_{gp}}) \\
\vdots & \ddots & \vdots \\
 k_{ii^{'}}(\bm{x}^{N_{gp}},\bm{x}^1) & \cdots & k_{ii^{'}}(\bm{x}^{N_{gp}},\bm{x}^{N_{gp}}) \\
\end{pmatrix} \\
& \\
\end{aligned},
\end{equation} $i,i^{'}=1,...,m$. $\bm{\Sigma}^{'}=\bm{\Sigma} \otimes I_{N_{gp}}\in\mathbb{R}^{mN_{gp}\times mN_{gp}}$ represents a diagonal noise matrix, where the diagonal elements of $\bm{\Sigma}$ are $\sigma^2_i(\bm{x}^{'}),i=1,..,m$. The performance of the MOGP mainly depends on the multi-output covariance matrix $\mathbf{K}_w(\bm{x},\bm{x}^{'})$. In recent years, using deep neural networks to learn kernel functions $\mathbf{K}_w$ has become a promising direction \cite{jakkala2021deep}, with the aim of improving the limited representation ability of traditional manually designed kernel functions.

\section{Problem Formulation}
\label{sec3}
Given a target dataset $\mathcal{D} = \{\mathcal{D}_{tra}, \mathcal{D}_{val}, \mathcal{D}_{tet}\}$, the multi-objective GNAS problem can be formulated as:
\begin{equation}\label{GNAS_eq}
\begin{split}
A^* \in \underset{\bm{\alpha} \in \mathcal{A}}{\operatorname{argmin}} & \ \bm{\mathcal{L}}(\bm{\alpha})=(l_1(\bm{\alpha}; \bm{w}^*),...,l_m(\bm{\alpha}; \bm{w}^*))^T\\
s.t. &\ \bm{w}^*(\bm{\alpha})\in \underset{\bm{w} \in \mathcal{W}}{\operatorname{argmin}} \ \mathcal{F}(\bm{w};\bm{\alpha})\\
\end{split},
\end{equation}
where $\bm{\alpha}$ and $\bm{w}$ denote the architecture and its associated weights, respectively. $\mathcal{A}$ and $\mathcal{W}$ are the architecture and weight spaces, respectively.
$\bm{\mathcal{L}}=(l_1,...,l_m)^T$ is the upper-level vector-valued function with $m$ objectives on the validation set $\mathcal{D}_{val}$, such as \#Acc and \#Params. $\mathcal{F}$ is the loss function on the training set $\mathcal{D}_{tra}$ for a candidate architecture $\bm{\alpha}$.

Supposing that $\bm{\alpha}_{1}$ and $\bm{\alpha}_{2}$ are two candidate architectures, $\bm{\alpha}_{1}$ is said to Pareto dominate $\bm{\alpha}_{2}$, if and only if $l_i(\bm{\alpha}_{1})\leq l_i(\bm{\alpha}_{2}),\forall i\in\{1, \cdots, m\}$ and there exists at least one objective $l_j (j\in\{1, \cdots, m\})$ satisfying $l_j(\bm{\alpha}_{1})< l_j(\bm{\alpha}_{2})$. $\bm{\alpha}^*$ is Pareto optimal if no $\bm{\alpha}$ exists such that $\bm{\alpha}$ dominates $\bm{\alpha}^*$. The set of all Pareto architectures is known as the Pareto set $A^*$. The projection of the Pareto set in the objective space is known as the Pareto front. In summary, multi-objective GNAS aims to determine an architecture set that approximates the Pareto set \cite{996017}. In lower-level optimization, learning the optimal weights for an architecture requires costly stochastic gradient descent over multiple epochs. Therefore, the bilevel optimization problem is computationally expensive.

We consider the multi-objective GNAS problem on a benchmark dataset (such as the Cora dataset) in NAS-Bench-Graph as a source task. KEGNAS aims to improve the solving efficiency of a GNAS method with the aid of the knowledge acquired from NAS-Bench-Graph, which is formally defined as:
\begin{equation}\label{KEGNAS_eq}
A^*=\underset{\bm{\alpha} \in \mathcal{A}}{\operatorname{argmin}} \ [\bm{\mathcal{L}}(\bm{\alpha})|\mathcal{M}],
\end{equation}
where $\mathcal{M}=\{\mathcal{A},\bm{\mathcal{L}}^k,\mathcal{D}_b^k,k=1,...,K\}$ is the prior database containing the available data from $K$ source tasks in NAS-Bench-Graph. $\mathcal{A}$ is the wealth of architecture libraries. For the $k$th source task, $\bm{\mathcal{L}}^k$ represents the performance metrics (\#Acc and \#Params) of the architectures in $\mathcal{A}$ on the benchmark dataset $\mathcal{D}_b^k$. 

\section{Proposed Framework}
\label{sec4}
Fig. \ref{overview} presents an overview of the KEGNAS framework, which consists of two phases: training and search. In the following, we introduce the knowledge model, DMOGP, construction of training data from NAS-Bench-Graph, and framework of KEGNAS in detail.

\begin{figure*}
\centering
\includegraphics[width=0.95\textwidth]{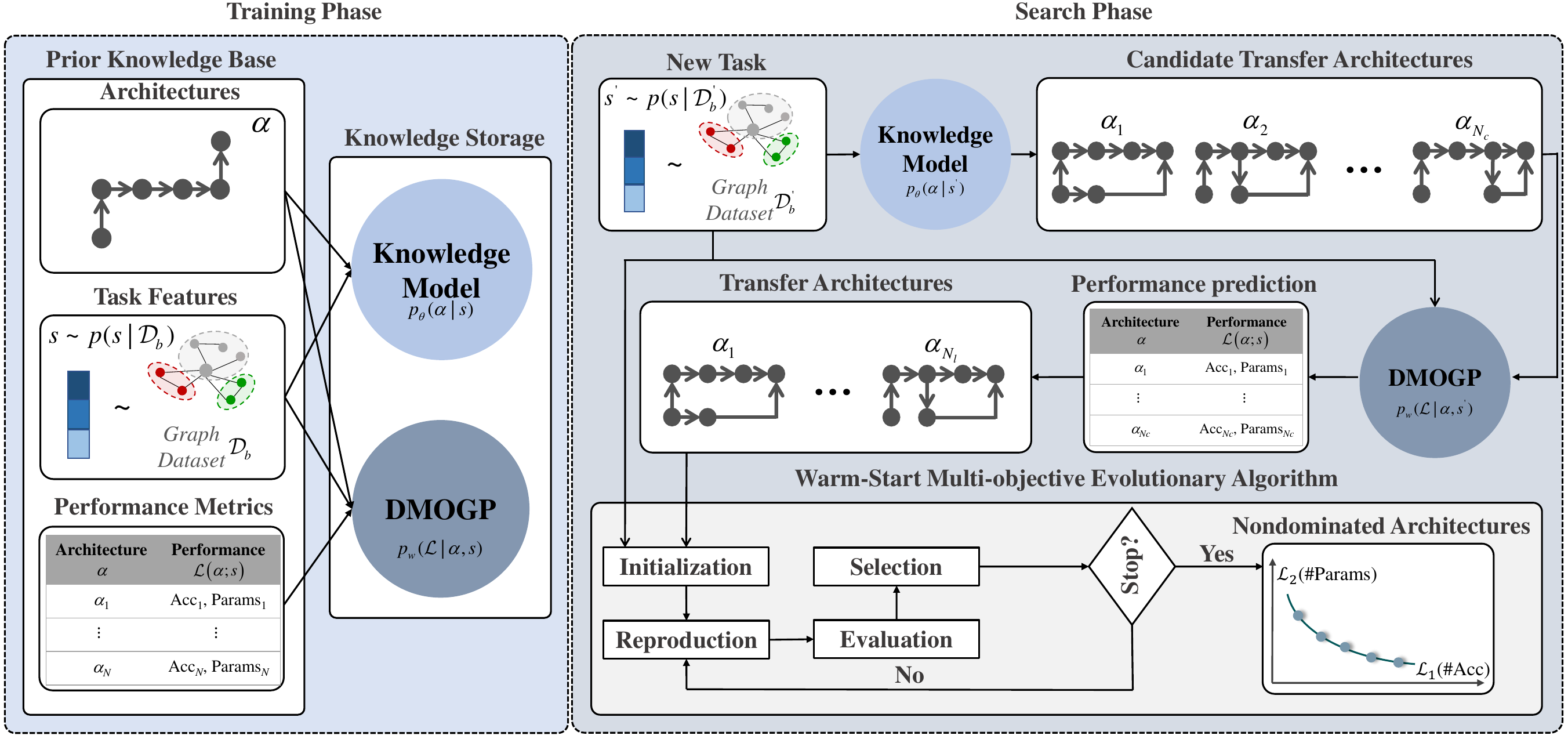}
\caption{Overview of KEGNAS, which consists of two phases: training and search. A knowledge model and a DMOGP learn how to generate and evaluate candidate transfer architectures for a new task based on a prior knowledge base from NAS-Bench-Graph, respectively. The selected transfer architectures are employed to warm-start the MOEA.} \label{overview}
\end{figure*}

\subsection{Knowledge Model}
In KEGNAS, a knowledge model learns the mapping from datasets to architectures, allowing for the rapid generation of candidate transfer architectures for a new NAS task on an unseen dataset. Training is performed on a prior knowledge base $\mathcal{M}$ containing $K$ source NAS tasks from NAS-Bench-Graph. This section describes the construction of the knowledge model. We assume that $K$ source NAS tasks obey a task family $\mathcal{T}(\bm{\alpha};\bm{s})$:
\begin{equation}\label{task_family}
\mathcal{T}(\bm{\alpha};\bm{s})=\{\bm{\mathcal{L}}(\bm{\alpha};\bm{s})|\bm{s} \in \mathcal{S}\}, \bm{\alpha} \in \mathcal{A},
\end{equation}
where $\mathcal{S}=\{\bm{s}^1,...,\bm{s}^K\}$ denotes a task space that contains $K$ source tasks and $\bm{s}$ is the task feature describing the source task in $\mathcal{T}$. In NAS-Bench-Graph, each NAS task corresponds to a unique benchmark dataset. The task feature $\bm{s}$ can be extracted from the benchmark dataset $\mathcal{D}_b$ using graph embedding methods. Assuming that the task features $\bm{s}$ follow a probability distribution $p(\bm{s}|\mathcal{D}_b)$, (\ref{task_family}) can be rewritten as:
\begin{equation}\label{task_features}
\begin{split}
    \mathcal{T}(\bm{\alpha};\bm{s})=&\{\bm{\mathcal{L}}(\bm{\alpha};\bm{s})|\bm{s} \sim p(\bm{s}|\mathcal{D}_b)\}, \\
    & \mathcal{D}_b\in\{\mathcal{D}^1_b,...,\mathcal{D}^K_b\} ,\bm{\alpha} \in \mathcal{A}.
\end{split}
\end{equation}

In this study, the commonly used graph embedding method, variational graph autoencoder (VGAE) \cite{kipf2016variational}, is adopted to obtain the task feature distribution $p_{\bm{\phi}}(\bm{s}|\mathcal{D}_b)$ parameterized by $\bm{\phi}$ directly. In VGAE, the encoder maps the nodes in a graph to a distribution in the latent space, while the decoder reconstructs the graph structure using the latent space representations. By optimizing the model parameters $\bm{\phi}$ based on the reconstruction loss and a regularization term, VGAE can learn the feature distribution $p_{\bm{\phi}}(\bm{s}|\mathcal{D}_b)=\mathcal{N}(\bm{\mu},\bm{\sigma}^2)$ on the benchmark dataset $\mathcal{D}_b$. For more detailed explanations of the VGAE, please refer to \cite{kipf2016variational}. 



Given a task feature distribution $p(\bm{s}|\mathcal{D}_b)$, the Pareto set $\bm{A}^{*}$ for the NAS task is formalized as follows:
\begin{equation}\label{task_unique}
\bm{A}^{*} = \underset{\bm{\alpha} \in \mathcal{A}}{\operatorname{min}} \ \bm{\mathcal{L}}(\bm{\alpha};\bm{s}\sim p(\bm{s}|\mathcal{D}_b)).
\end{equation}

\begin{figure}[t]
\centering
\includegraphics[width=0.95\textwidth]{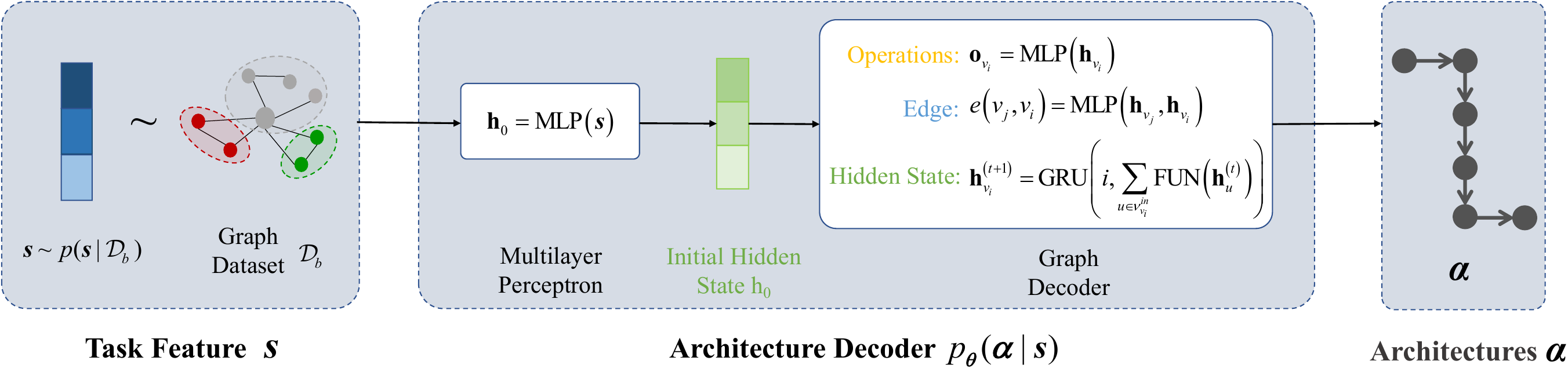}
\caption{Overview of the knowledge model.} \label{knowledge_model}
\end{figure}

We consider building a knowledge model to learn the mapping between the task feature distribution $p(\bm{s}|\mathcal{D}_b)$ and the Pareto architecture $\bm{\alpha}^*\in\bm{A}^*$ directly using the prior knowledge base $\mathcal{M}$ from NAS-Bench-Graph. \textcolor{black}{An overview of the knowledge model is presented in Fig. \ref{knowledge_model}.} For each NAS task, the knowledge model can generate a new architecture $\bm{\alpha}$ from the task feature $\bm{s}\sim p(\bm{s}|\mathcal{D}_b)$ using the architecture decoder $p_{\bm{\theta}}(\bm{\alpha}|\bm{s})$. \textcolor{black}{Subsequently, the architecture decoder $p_{\bm{\theta}}(\bm{\alpha}|\bm{s})$ can be learned by minimizing the cross-entropy as follows:}
\begin{equation}\label{ELBO}
\underset{\bm{\theta}}{\operatorname{min}} \ \mathbb{E}_{\bm{s}\sim p_{\bm{\phi}}(\bm{s}|\mathcal{D}_b)}[-\log p_{\bm{\theta}}(\bm{\alpha}|\bm{s})].
\end{equation}

Because each architecture is encoded as a DAG, we directly adopt the existing popular graph decoder for DAGs \cite{lee2021rapid,NEURIPS2019_e205ee2a} as the architecture decoder $p_{\bm{\theta}}(\bm{\alpha}|\bm{s})$, which only passes messages along the topological order. 
\textcolor{black}{As shown in Fig. \ref{knowledge_model}, given a task feature $\bm{s} \sim p(\bm{s}|\mathcal{D}_b)$, a multi-layer
perceptron (MLP) followed by Tanh is used to map $\bm{s}$ to an initial hidden state $\mathbf{h}_{0}$. The graph decoder then constructs the DAG node by node. Specifically, constructing the $i$th node $v_i$ according to the topological order involves two steps: 1) determining the operation type $\mathbf{o}_{v_i} \in \mathbb{R}^{1 \times n}$ of node $v_i$ over $n$ operations and 2) predicting the existence of edges between node $v_i$ and all previously processed nodes $\{v_j | j = i-1, \ldots, 1\}$.}

\textcolor{black}{First, an MLP with Softmax is used to obtain the probability distribution $\mathbf{o}_{v_i} \in \mathbb{R}^{1 \times n}$ of operation types for node $v_i$ based on the current hidden state $\mathbf{h}_{v_i}$, i.e. $\mathbf{o}_{v_i} =\text{MLP}({\mathbf{h}_{v_i}})$. When $v_i$ is identified as a graph termination node, the construction process terminates and all leaf nodes are connected to $v_i$. Otherwise, the hidden state at time step $t+1$ is updated as follows:}

\begin{equation}\label{kn_update}
\textcolor{black}{
\mathbf{h}_{v_i}^{(t+1)}=\text{GRU}(i,\sum_{u\in \mathcal{V}_{v_i}^{in}}\text{FUN}(\mathbf{h}_u^{(t)})),}
\end{equation}
\textcolor{black}{where $\mathcal{V}_{v_i}^{in}$ represents the set of predecessor nodes with incoming edges to $v_i$. $\text{GRU}$ denotes a gated recurrent unit. $\text{FUN}$ comprises the mapping and gating functions with MLPs. Then, an MLP with Sigmoid is employed to compute the edge connection probability $e(v_j, v_i)$ between each previously processed node $\{v_j | j = i-1, \ldots, 1\}$ and $v_i$ based on the hidden states $\mathbf{h}_{v_j}$ and $\mathbf{h}_{v_i}$. In particular, $e(v_j, v_i)=\text{MLP}(\mathbf{h}_{v_j},\mathbf{h}_{v_i})$. Finally, whenever a new edge is connected to $v_i$, the hidden state $\mathbf{h}_{v_i}$ is updated using the formula (\ref{kn_update}). During testing, edges with $e(v_j, v_i)>0.5$ and operations with maximum probabilities are selected for the generated architecture. A detailed description of the graph decoder can be found in \cite{NEURIPS2019_e205ee2a,pmlr-v198-zhu22a}.} 

During the testing phase, the trained knowledge model can rapidly generate high-quality candidate transfer architectures for a new task based on its task feature distribution.

\subsection{DMOGP}
According to the formula (\ref{GNAS_eq}), the evaluation process of each architecture must complete a lower-level optimization that involves multiple costly iterations of gradient descent over multiple epochs. \textcolor{black}{Therefore, directly evaluating a large number of candidate transfer architectures in the target task to select a transfer architecture is computationally expensive.} We construct a surrogate model to reduce the computational cost of the transfer architecture selection. Given a dataset $\mathcal{D}_{gp}=\{ \bm{\alpha}^n,\bm{\mathcal{L}}^n,\bm{s}\sim p(\bm{s}|\mathcal{D}_b^n),n=1,...,N_{gp}\}$, we employ a DMOGP to fit the functional relationship between architectures $\bm{\alpha}$ and multiple performance metrics $\bm{\mathcal{L}}$ with the task feature $\bm{s}$.

The flowchart of the DMOGP is shown in Fig. \ref{fig2}. First, We use a graph encoder $\bm{f}$ to extract the features of the architecture $\bm{\alpha}$. Subsequently, the architecture features $\bm{f}(\bm{\alpha})$ and task features $\bm{s}$ are post-processed by a fully connected neural network $\bm{g}$, whose output is encoded into the kernel of the MOGP. The construction procedure described above results in the surrogate model being learnable end-to-end. Based on task features and architecture features, the DMOGP can be used to predict the performance of candidate transfer architectures for different tasks. We provide the details of the different components in the following.

\begin{figure}[t]
\centering
\includegraphics[width=0.75\textwidth]{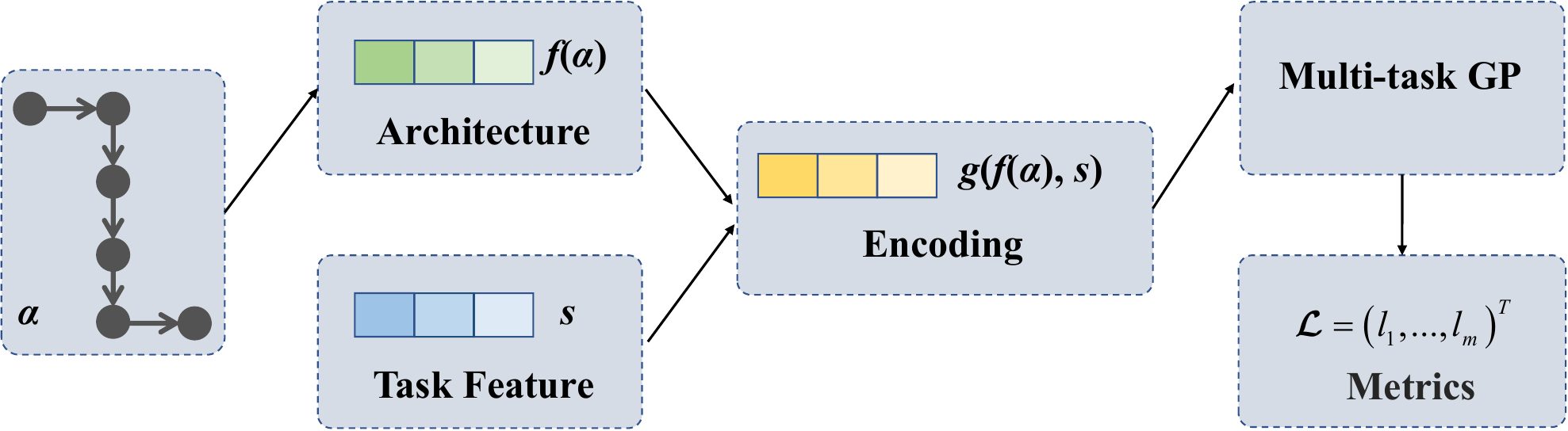}
\caption{Flowchart of the DMOGP.} \label{fig2}
\end{figure}

\paragraph{Architecture Features} A graph encoder $\bm{f}_{\bm{w}_1}$ for DAGs is adopted to extract the architecture features \cite{NEURIPS2019_e205ee2a}. The graph encoder performs message passing along the topological order via GRUs to encode the graph architecture. 
\textcolor{black}{One GRU unit traverses the topological order of the DAG from input to output, and another traverses the DAG in reverse topological order. \textcolor{black}{A latent graph representation can be obtained through this process.} Subsequently, an MLP is employed to map the latent graph representation to the architecture feature.} Details of graph encoder architectures can be found in \cite{shala2023transfer,NEURIPS2019_e205ee2a}.

\paragraph{Encoding for the MOGP Kernel} The simplest fully connected neural network with $c$ hidden neurons $\bm{g}_{\bm{w}_2}:\mathbb{R}^{a+b}\rightarrow{\mathbb{R}^{c}}$ is constructed to post-process the task features $\bm{s}\in \mathbb{R}^{a}$ and architecture features $\bm{f}_{\bm{w}_1}(\bm{\alpha})\in \mathbb{R}^{b}$. Its output $\bm{g}_{\bm{w}_2}(\bm{f}_{\bm{w}_1}(\bm{\alpha}),\bm{s})$ is viewed as the 
 encoding of the kernel of the MOGP.

\paragraph{MOGP} For any two samples $\{ \bm{\alpha}^i,s^i\}$ and $\{ \bm{\alpha}^j,s^j\}$ in the dataset $\mathcal{D}_{gp}$, the deep kernel function of the MOGP can be formalized as:
\begin{equation}\label{MOGP} 
\begin{aligned}
    &\mathbf{K}_{\bm{w}_3}\left((\bm{\alpha}^i,s^i),(\bm{\alpha}^j,s^j)\right)=\\
    & \mathbf{K}_{\bm{w}_3}\left(\bm{g}_{\bm{w}_2}(\bm{f}_{\bm{w}_1}(\bm{\alpha}^i),\bm{s}^i),\bm{g}_{\bm{w}_2}(\bm{f}_{\bm{w}_1}(\bm{\alpha}^j),\bm{s}^j)\right) \\
\end{aligned},
\end{equation}
where $\bm{w}_3$ denotes the parameters of the deep kernel function. Therefore, the learnable parameters for the entire DMOGP are $\bm{w} = (\bm{w}_1,\bm{w}_2,\bm{w}_3)$. Subsequently, the parameters $\bm{w}$ can be optimized by maximizing the log marginal likelihood function
\begin{equation}\label{MOGP_ml} 
\underset{\bm{w}}{\operatorname{max}} \ \log p_{\bm{w}}(\mathcal{L}|\bm{\alpha},\bm{s}).
\end{equation}

\subsection{Construction of Training Data from NAS-Bench-Graph}

\begin{algorithm} [htbp]
\caption{Construction of training data}
\label{alg1}
\begin{algorithmic}[1]
\Require $K$: Number of source tasks; $N_s$: Number of non-dominated fronts for each source task; $\mathcal{M}=\{\mathcal{A},\bm{\mathcal{L}}^k,\mathcal{D}_b^k,k=1,...,K\}$: Prior knowledge base from NAS-Bench-Graph.
\State $\mathcal{D}_{km} \leftarrow \emptyset,N_{km} \leftarrow 0$;
\State $\mathcal{D}_{gp} \leftarrow \emptyset,N_{gp} \leftarrow 0$;
\For{$k = 1$ to $K$}
  \State $p(\bm{s}|\mathcal{D}_b^k) \leftarrow$ Extract task feature distribution from $\mathcal{D}_b^k$ via VGAE;
  \State $\mathcal{A}_{nd} \leftarrow$ \textit{Fast-Nondominated-Sorting}($\mathcal{A}, \bm{\mathcal{L}}^k, N_s$);
  \ForEach{$\bm{\alpha}\in\mathcal{A}_{nd}$}
  \State $\mathcal{D}_{km} \leftarrow \mathcal{D}_{km} \cup \{\bm{\alpha},\bm{s}\sim p(\bm{s}|\mathcal{D}_b^k)\}$;
  \State $\mathcal{D}_{gp} \leftarrow \mathcal{D}_{gp} \cup \{\bm{\alpha},\bm{\mathcal{L}}^k(\bm{\alpha}),\bm{s}\sim p(\bm{s}|\mathcal{D}_b^k)\}$;
  \State $N_{km} \leftarrow N_{km} + 1, N_{gp} \leftarrow N_{gp} + 1$;
  \EndFor
\EndFor
\Ensure $\mathcal{D}_{km}=\{ \bm{\alpha}^n,\bm{s}\sim p(\bm{s}|\mathcal{D}_b^n),n=1,...,N_{km}\}$: Training data for knowledge model; $\mathcal{D}_{gp}=\{ \bm{\alpha}^n,\bm{\mathcal{L}}^n,\bm{s}\sim p(\bm{s}|\mathcal{D}_b^n),n=1,...,N_{gp}\}$: Training data for DMOGP.
\end{algorithmic}
\end{algorithm}

As shown in Algorithm \ref{alg1}, we introduce obtaining the task feature distribution and Pareto sets from the prior knowledge base (NAS-Bench-Graph) $\mathcal{M}=\{\mathcal{A},\bm{\mathcal{L}}^k,\mathcal{D}_b^k,k=1,...,K\}$ to compose the training data ($\mathcal{D}_{km}$ and $\mathcal{D}_{gp}$) of the knowledge model and DMOGP. For the $k$th source task, we first extract the task feature distribution $p(\bm{s}|\mathcal{D}_b^k)$ from benchmark datasets $\mathcal{D}_b^k$ via VGAE \cite{kipf2016variational} (Line 4 in Algorithm \ref{alg1}). We then perform \textit{Fast-Nondominated-sorting} \cite{996017} on all architectures in the architecture library $\mathcal{A}$ according to the performance metrics $\bm{\mathcal{L}}^k$ (Line 5 in Algorithm \ref{alg1}). Each architecture $\bm{\alpha}$ in the top $N_s$ non-dominated fronts, performance metrics $\bm{\mathcal{L}}^k(\bm{\alpha})$, and task feature $\bm{s}$ are combined to form a training sample (Lines 6-10 in Algorithm \ref{alg1}). Then, $\mathcal{D}_{km}$ and $\mathcal{D}_{gp}$ are employed to train the knowledge model and DMOGP, respectively.

\textcolor{black}{NAS-Bench-Graph provides multiple performance metrics for all architectures in the designed search space on many benchmark datasets. Each graph classification task in the benchmark dataset can be viewed as a source task. The training data for the knowledge model and DMOGP comprise numerous non-dominated fronts gathered from the entire search space on multiple source tasks, ensuring high quality and diversity.}

\subsection{Framework of KEGNAS}

\begin{algorithm} [htbp]
\caption{Framework of KEGNAS}
\label{alg2}
\begin{algorithmic}[1]
\Require $\mathcal{M}=\{\mathcal{A},\bm{\mathcal{L}}^k,\mathcal{D}_b^k,k=1,...,K\}$: Prior knowledge base from NAS-Bench-Graph; $\mathcal{D}_n$: Datasets of the new task; $N_c$: Number of candidate transfer architectures; $N_p$: Population size; $G$: Maximum number of generations.
\State $\left(\mathcal{D}_{km},\mathcal{D}_{gp}\right) \leftarrow $ Obtain the training data using Algorithm \ref{alg1};
\State $p_{\bm{\theta}}(\bm{\alpha}|\bm{s}) \leftarrow$ Train a knowledge model on $\mathcal{D}_{km}$ by minimizing (\ref{ELBO});
\State $p_{\bm{w}}(\mathcal{L}|\bm{\alpha},\bm{s}) \leftarrow$ Train a DMOGP on $\mathcal{D}_{gp}$ by maximizing (\ref{MOGP_ml});
\State $\bm{s}^{'}\sim p(\bm{s}^{'}|\mathcal{D}_n) \leftarrow$ Extract the task feature distribution from $\mathcal{D}_n$ via VGAE;
\State $\bm{A}^{'}=\{\bm{\alpha}_1,...,\bm{\alpha}_{N_c}\} \leftarrow $ Generate candidate transfer architectures for a new task with feature $\bm{s}^{'}$ via $p_{\bm{\theta}}(\bm{\alpha}|\bm{s}^{'})$;
\State $\bm{\mathcal{L}}^{'}=\{\bm{\mathcal{L}}(\bm{\alpha}_1),...,\bm{\mathcal{L}}(\bm{\alpha}_{N_c})\} \leftarrow $ Evaluate the candidate transfer architectures on a new task with feature $\bm{s}^{'}$ via $p_{\bm{w}}(\mathcal{L}|\bm{\alpha},\bm{s}^{'})$;
\State $\bm{A}^{''}=\{\bm{\alpha}_1,...,\bm{\alpha}_{N_l}\} \leftarrow$ Select transfer architectures using \textit{Fast-Nondominated-Sorting} ($\bm{A}^{'}, \bm{\mathcal{L}}^{'}, 1$);
\If{$N_l < N_p$} // Population Initialization
    \State $\bm{P}_{\bm{A}} \leftarrow$ Randomly initialize $N_p-N_l$ architectures for a new task with feature $\bm{s}^{'}$;
    \State $\bm{P}_{\bm{A}} \leftarrow \bm{P}_{\bm{A}} \cup \bm{A}^{''}$;
\Else
    \State $\bm{P}_{\bm{A}} \leftarrow$ Randomly select $N_p$ architectures from $\bm{A}^{''}$;
\EndIf

\For{$g = 1$ to $G$}
  \State $\bm{P}_{\bm{A}} \leftarrow$ Update population by conducting the MOEA;
\EndFor
\Ensure $\bm{P}_{\bm{A}}^*$: Non-dominated architectures in $\bm{P}_{\bm{A}}$.
\end{algorithmic}
\end{algorithm}

Algorithm \ref{alg2} presents the framework of KEGNAS. KEGNAS consists of two stages: training and search. In the training phase, we first obtain the training data from NAS-Bench-Graph, including the architecture library, task features, and performance metrics (Algorithm \ref{alg1}). Subsequently, a knowledge model and DMOGP are trained to store this prior knowledge (Lines 2-3 in Algorithm \ref{alg2}). In the search phase, the knowledge model is employed to generate a set of candidate transfer architectures for a new task (Line 5 in Algorithm \ref{alg2}). Then, the DMOGP model is used to evaluate the performance of the candidate transfer architectures on the new tasks (Line 6 in Algorithm \ref{alg2}). Non-dominated candidate transfer architectures are selected to initialize the population (Line 7 in Algorithm \ref{alg2}). As shown in Fig. \ref{fig1}, each architecture in the population is encoded as an eight-bit vector. When the number of transfer architectures is less than the predefined population size, the remaining architectures are randomly sampled from the search space of NAS-Bench-Graph; otherwise, $N_p$ architectures are randomly selected from the transfer architectures as the initial population (Lines 8-13 in Algorithm \ref{alg2}). We can then employ any MOEA to update the population, such as NSGAII \cite{996017} (Lines 14-16 in Algorithm \ref{alg2}). Finally, we obtain non-dominated architectures from the final population for decision-makers to choose from. In KEGNAS, the training phase must be executed only once to store prior knowledge. For a new task, we only need to perform the search phase (Lines 4-16 in Algorithm \ref{alg2}). In the search phase, the trained knowledge
model and DMOGP can quickly generate and evaluate candidate transfer architectures with a low computational cost to warm-start the MOEA.

\textcolor{black}{The knowledge model and DMOGP are exclusively employed for initializing the population (Lines 4-13 in Algorithm \ref{alg2}), rather than being utilized within the components of the MOEA (Lines 14-16 in Algorithm \ref{alg2}), such as evaluation, selection, and reproduction. This design endows KEGNAS with high flexibility. Specifically, prior knowledge is efficiently infused into the initial population using transfer architectures as carriers, without requiring any prior assumptions or modifications to the embedded MOEA. \textcolor{black}{Consequently, in practice, KEGNAS can be seamlessly integrated with any MOEA designed to solve GNAS problems without disrupting or interfering with existing evaluation strategies that are crucial for reducing computational costs.}}

\section{Experimental Studies}
\label{sec5}
This section presents a series of experimental studies on NAS-Bench-Graph and five real-world graph datasets to validate the KEGNAS framework. In addition, the effectiveness of prior knowledge utilization in KEGNAS is further explored. All experiments are running with PyTorch (v. 1.10.1) and PyTorch Geometric\footnote{\url{https://github.com/pyg-team/pytorch\_geometric}\label{pyg}} (v. 2.0.3) on a GPU 2080Ti (Memory: 12GB, Cuda version:11.3).

\subsection{Experimental Studies on NAS-Bench-Graph}

\subsubsection{Experimental Settings}

\begin{table}[htbp]
\scriptsize
\centering
\caption{Details of datasets in NAS-Bench-Graph.}\label{datasets}
\begin{tabular}{ccccc}
\toprule
Dataset & \#Nodes & \#Edges & \#Features & \#Classes \\
\midrule
Cora & 2,708 & 5,429 & 1,433 & 7 \\
CiteSeer & 3,327 & 4,732 & 3,703 & 6 \\
PubMed & 19,717 & 44,338 & 500 & 3 \\
CS & 18,333 & 81,894 & 6,805 & 15 \\
Physics & 34,493 & 247,962 & 8,415 & 5\\
Photo & 7,487 & 119,043 & 745 & 8 \\
Computers & 13,381 & 245,778 & 767 & 10 \\
ArXiv & 169,343 & 1,166,243 & 128 & 40 \\
\bottomrule
\end{tabular}
\end{table}

\paragraph{Datasets and Tasks} Owing to the uniqueness of the search space of ogbn-proteins, we only adopt eight public datasets in NAS-Bench-Graph, with the exception of ogbn-proteins. \textcolor{black}{The statistics of these datasets, which cover various types of complex networks, are presented in Table \ref{datasets}.} Cora, CiteSeer, and PubMed \footnote{\url{https://github.com/kimiyoung/planetoid}} are citation networks in which each node is a paper and each edge is the citation relationship between two papers. CS and Physics \footnote{\url{https://github.com/shchur/gnn-benchmark/}\label{gnn_b}} are co-authorship networks from Microsoft. Photo and Computers \textsuperscript{\ref {gnn_b}} are co-purchase networks of Amazon. ArXiv \footnote{\url{https://ogb.stanford.edu/docs/nodeprop/}} is a citation network from the Open Graph Benchmark that represents the citation relationships between papers from arXiv. A more detailed description is provided in \cite{qin2022nasbenchgraph}.

\begin{figure*} [htbp]
	\centering
	\subfloat[Pearson correlation coefficient of \#Acc]{
		\includegraphics[width=0.33\linewidth]{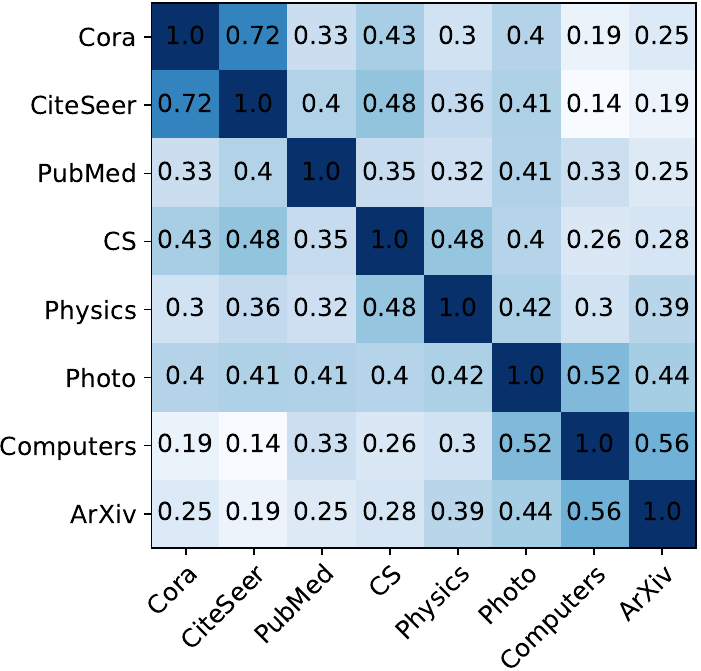}}
	\subfloat[Pearson correlation coefficient of \#Params]{
		\includegraphics[width=0.33\linewidth]{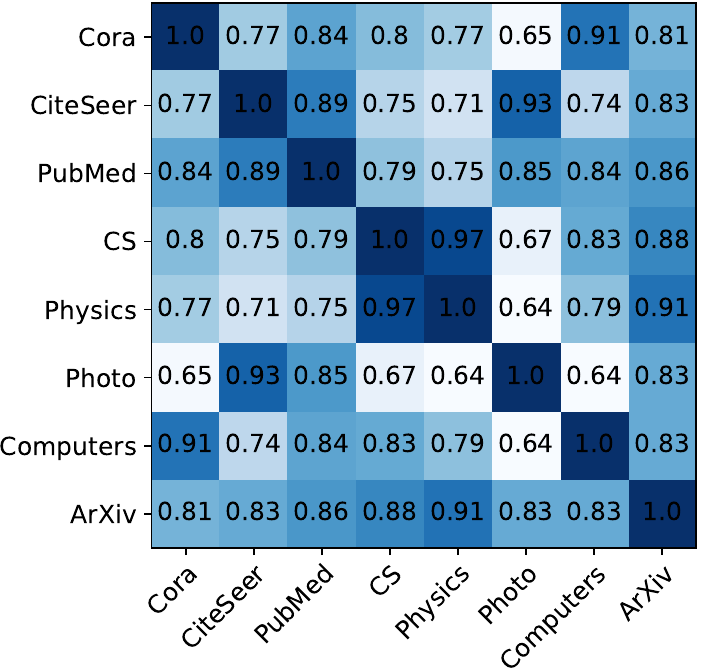}}
	\subfloat[Overlapping ratio of Pareto architecture set]{
		\includegraphics[width=0.33\linewidth]{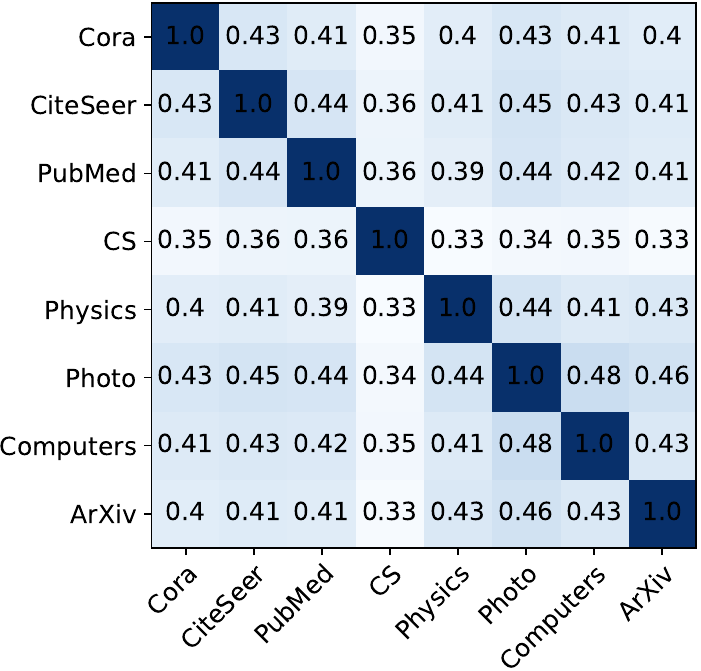}}
	\caption{Similarity across different tasks using three metrics.}
	\label{fig_sim}
\end{figure*}

As shown in Fig. \ref{fig_sim}, we adopt the three metrics to measure the similarity between tasks including the Pearson correlation coefficient of \#Acc, the Pearson correlation coefficient of \#Params, and the overlapping ratio of the Pareto architecture set. \textcolor{black}{It can be observed that a large gap existed in the similarity between different tasks}, which requires good transferability of GNAS methods. In addition, the overlapping ratio of the Pareto architecture set is relatively small, which further indicates that the top-performing architectures in the different graph datasets exhibit complex patterns.

For GNAS tasks on a dataset, the GNAS tasks on the remaining datasets in NAS-Bench-Graph are taken as source tasks. For example, for the GNAS task on the Cora dataset, the GNAS tasks on the other seven datasets are regarded as seven source tasks, thus constituting the prior database $\mathcal{M}$. \textcolor{black}{This experimental configuration guarantees that the prior knowledge base does not contain any information on the target GNAS task. In practical scenarios, for a new GNAS task, we can consider all GNAS tasks in NAS-Bench-Graph as source tasks, as discussed in the next section on experimental studies on real-world graph datasets.}
When constructing the training data from $\mathcal{M}$ (Algorithm \ref{alg1}), the number of non-dominated fronts $N_s$ for each source task is set to 10 to obtain samples containing rich knowledge.

\paragraph{Baselines} NSGAII \cite{996017}, a popular MOEA, is embedded in KEGNAS to form KEGNAS-NSGAII. The SBX crossover and PM mutation are used in NSGAII. It is recommended that the distribution indices of SBX and PM be set to 20 \cite{996017}. The probabilities of SBX and PM are set to 1 and $1/D$, respectively \cite{996017}, where $D$ is the dimension of the architecture encoding vector. To verify the effectiveness of KEGNAS, we provide six common baselines, namely GraphNAS \cite{ijcai2020p195}, Auto-GNN \cite{zhou2019auto}, Random \cite{pmlr-v115-li20c}, EA \cite{Real_Aggarwal_Huang_Le_2019}, RL \cite{zoph2017neural}, and NSGAII \cite{996017}. \textcolor{black}{The performance of the top 5\% architecture in NAS-Bench-Graph is also presented as a reference.}

GraphNAS and Auto-GNN are the most popular GNAS baselines, both of which automatically design an optimal neural network architecture based on RL. In contrast to GraphNAS, Auto-GNN employs a parameter-sharing strategy to improve the search efficiency, enabling homogeneous architectures to share parameters during training. Random, EA, and RL are the most classic NAS baselines for CNNs. As NAS-Bench-Graph provides graph architecture encoding and performance metrics, these baselines can be used directly to solve GNAS problems. The Random randomly samples an architecture each time. The EA evolves a population of architectures to approximate the optimal architecture. The RL trains a recurrent neural network using policy-based RL to generate high-quality architectures. Furthermore, because the search space of NAS-Bench-Graph contains most of state-of-the-art handcrafted architectures (such as ARMA\cite{9336270} and JK-Net\cite{pmlr-v80-xu18c}), handcrafted architectures are not employed as baselines alone.

\paragraph{Implementation Details} We implement all baselines using three existing open libraries, including AutoGL \footnote{\url{https://github.com/THUMNLab/AutoGL}}, NNI \footnote{\url{https://github.com/microsoft/nni/}}, and Geatpy \footnote{\url{https://github.com/geatpy-dev/geatpy}}. GraphNAS and Auto-GNN are implemented in AutoGL (v. 0.3.1). Random, EA, and RL are run through NNI (v. 2.6.1). Subsequently, NSGAII and KEGNAS-NSGAII are implemented using Geatpy (v. 2.7.0). To ensure a fair comparison, similar to the existing literature \cite{qin2022nasbenchgraph}, we only let each baseline access the performance metrics of 2\% of the architectures in the search space, that is, the maximum number of evaluations is $2\%\times26206 \approx 525$. For all baselines, the number of independent runs is set to 20.

For KEGNAS-NSGAII, we use the knowledge model to generate 500 candidate transfer architectures, that is, $N_c=500$. The population size $N_p$ is set to 25. For the training phase, all settings follow the previous NAS methods \cite{lee2021rapid,shala2023transfer}. Specifically, we train the knowledge model (and the DMOGP) using the standard Adam optimizer with a learning rate of 0.001 and a batch size of 256 for 400 epochs. In addition, the dimension of the task features $\bm{s}$ is set to 32.

\begin{sidewaystable}[hbtp]
\begin{center}
\scriptsize
\caption{Experimental results of all baselines on NAS-Bench-Graph in terms of \#Acc (\%).}
\label{Acc_Nas_Bench_Graph}
\begin{tabular}{ccccccccc|c}
\toprule
Baseline & Cora & CiteSeer & PubMed & CS & Physics & Photo & Computers & ArXiv & AR \\
\midrule
        GraphNAS & 81.97(0.17)  & 70.98(0.16) & 77.80(0.01)  & 90.96(0.06) & 92.43(0.02) & 92.46(0.02) & 84.75(0.19) & 71.88(0.03) & 4.63  \\
        Auto-GNN & 81.88(0.01)  & 70.77(0.12)  & 77.71(0.16)  & 91.01(0.05)  & 92.53(0.17)  & 92.46(0.03)  & 84.54(0.15)  & 72.13(0.02)  & 4.88  \\
        Random & 82.06(0.09)  & 70.48(0.07)  & 77.89(0.07)  & 90.87(0.08)  & 92.35(0.06)  & 92.49(0.02)  & 84.82(0.15)  & 71.99(0.07)  & 4.75  \\ 
        EA & 81.81(0.19)  & 70.57(0.12) & 77.93(0.11)  & 90.56(0.05)  & 92.33(0.07)  & 92.48(0.02)  & 84.31(0.29)  & 71.90(0.07)  & 6.00  \\ 
        RL & 82.18(0.22)  & 70.74(0.13)  & 77.96(0.10)  & 90.91(0.01)  & 92.56(0.04)  & 92.39(0.05)  & 84.91(0.19)  & 72.14(0.06)  & 3.25  \\ 
        NSGAII & 81.98(0.26)  & 70.91(0.27)  & 77.96(0.34)  & 90.88(0.20)  & 92.53(0.19)  & 92.50(0.15)  & 84.63(0.60)  & 72.13(0.09)  & 3.50  \\ 
        KEGNAS-NSGAII & \textbf{82.44(0.23)}  & \textbf{71.05(0.22)}  & \textbf{78.24(0.36)}  & \textbf{91.18(0.05)}  & \textbf{92.74(0.13)}  & \textbf{92.64(0.11)}  & \textbf{85.04(0.50)}  & \textbf{72.22(0.17)}  & 1.00 \\
        \midrule
        \textcolor{black}{Top 5\%} & \textcolor{black}{80.63} & \textcolor{black}{69.07}  & \textcolor{black}{76.60}  &  \textcolor{black}{90.01}  &  \textcolor{black}{91.67}  & \textcolor{black}{91.57}  &  \textcolor{black}{82.77}  &  \textcolor{black}{71.69}  & \\
\bottomrule
\end{tabular}
\end{center}
\end{sidewaystable}

\subsubsection{Experimental Results}

\begin{figure*}[htbp]
	\centering
	\subfloat[Cora]{
		\includegraphics[width=0.36\linewidth]{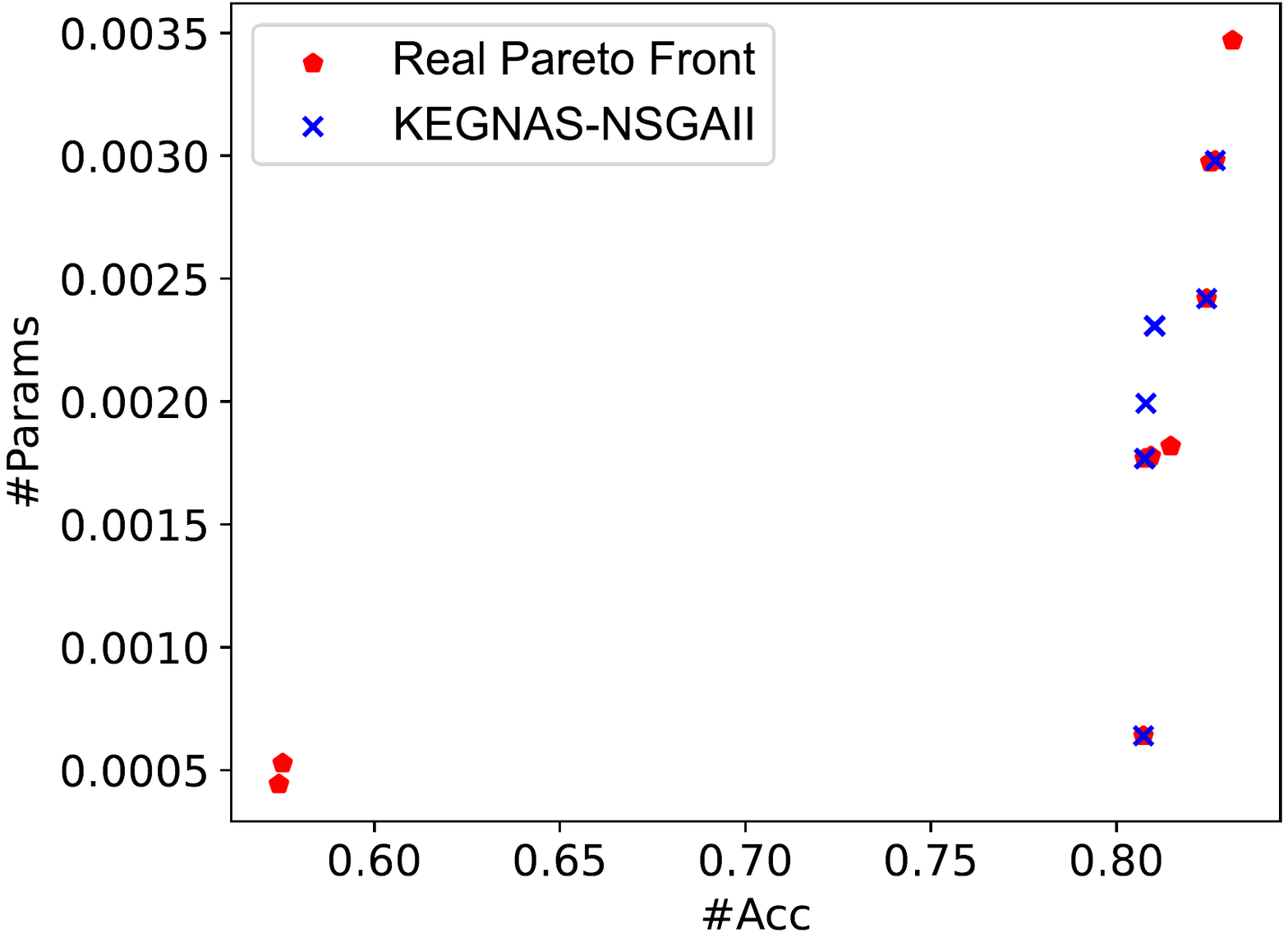}}
	\quad
	\subfloat[CiteSeer]{
		\includegraphics[width=0.36\linewidth]{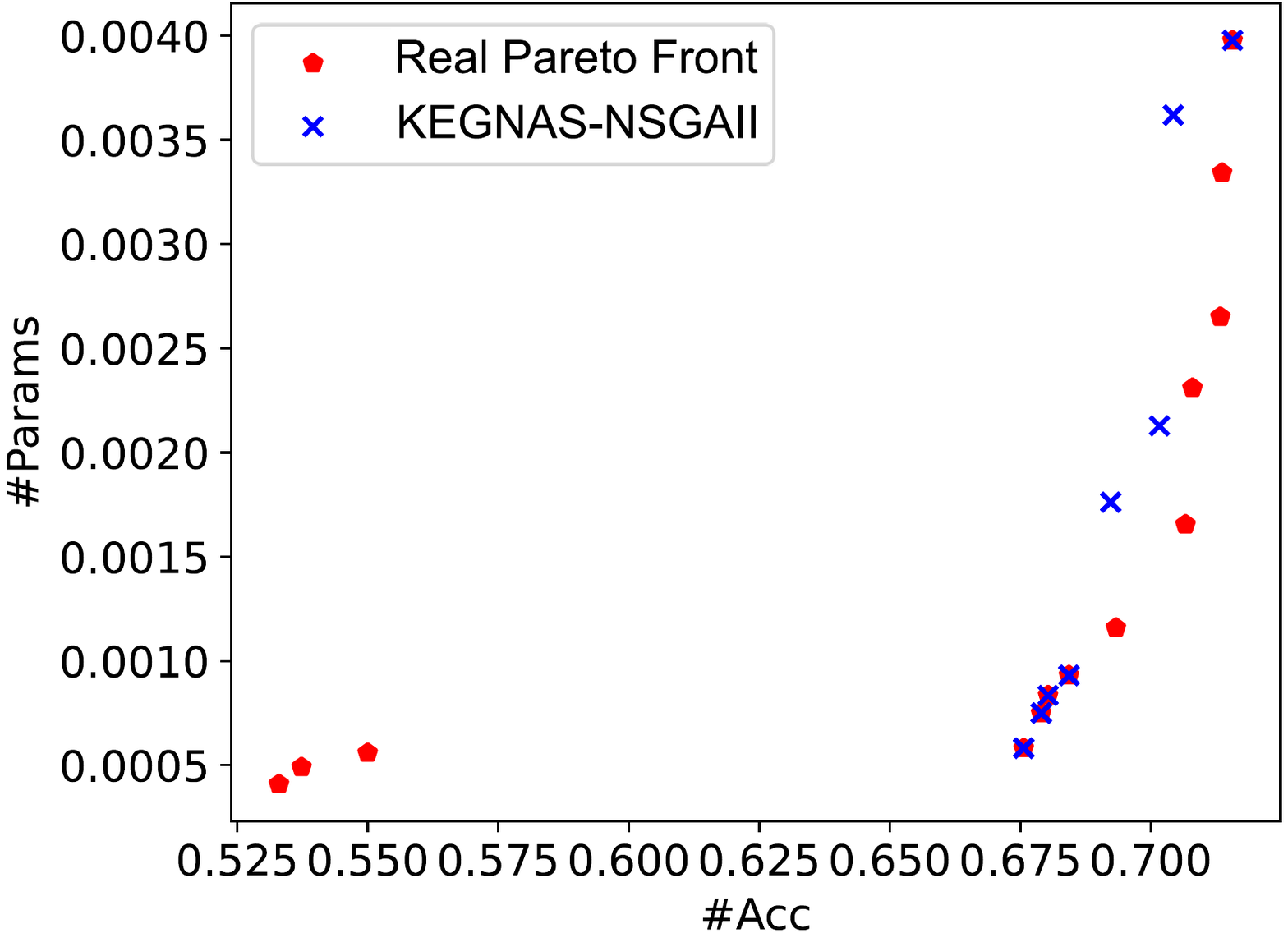}}
	\quad
	\subfloat[PubMed]{
		\includegraphics[width=0.36\linewidth]{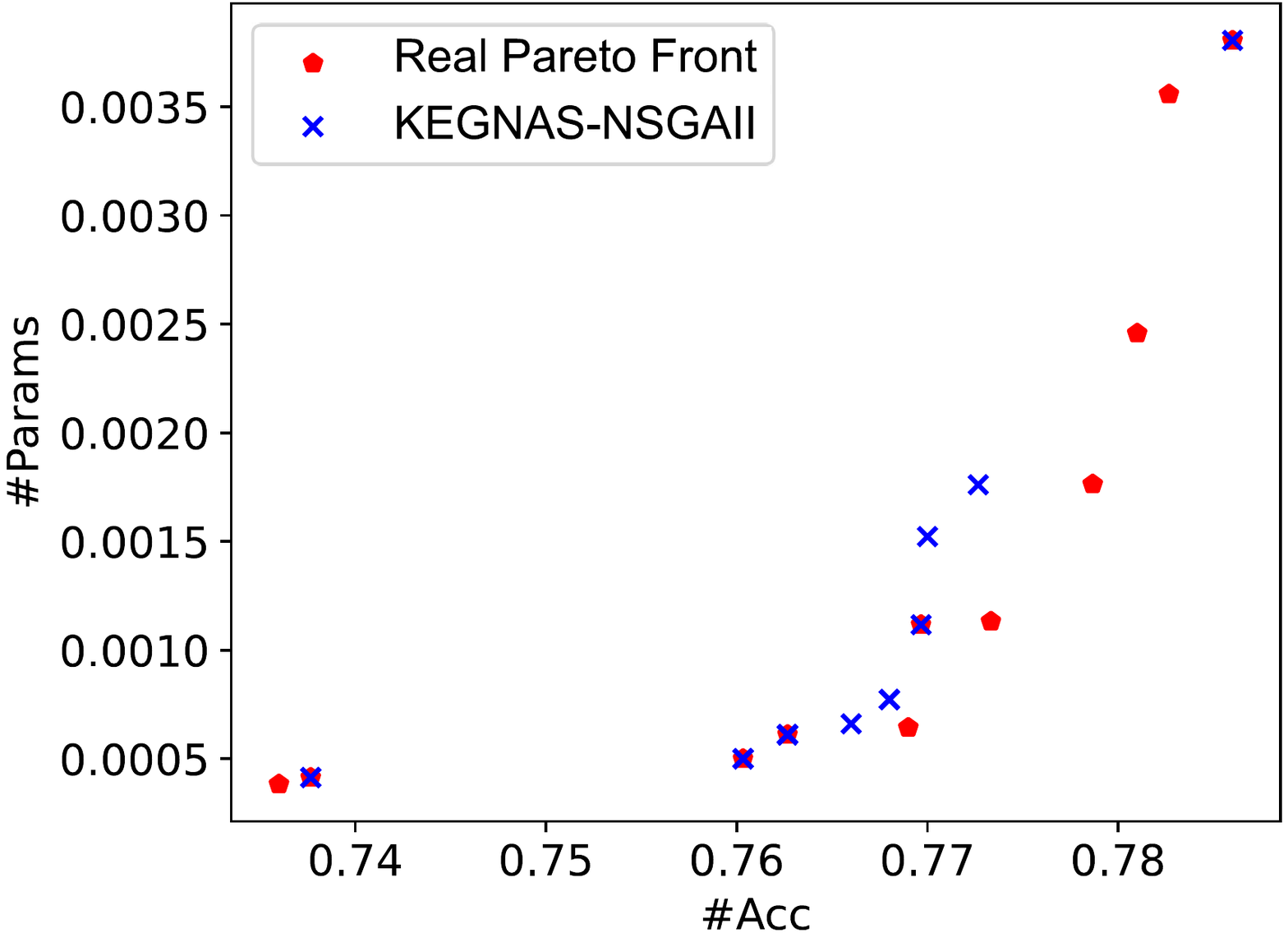}}
	\quad
	\subfloat[CS]{
		\includegraphics[width=0.36\linewidth]{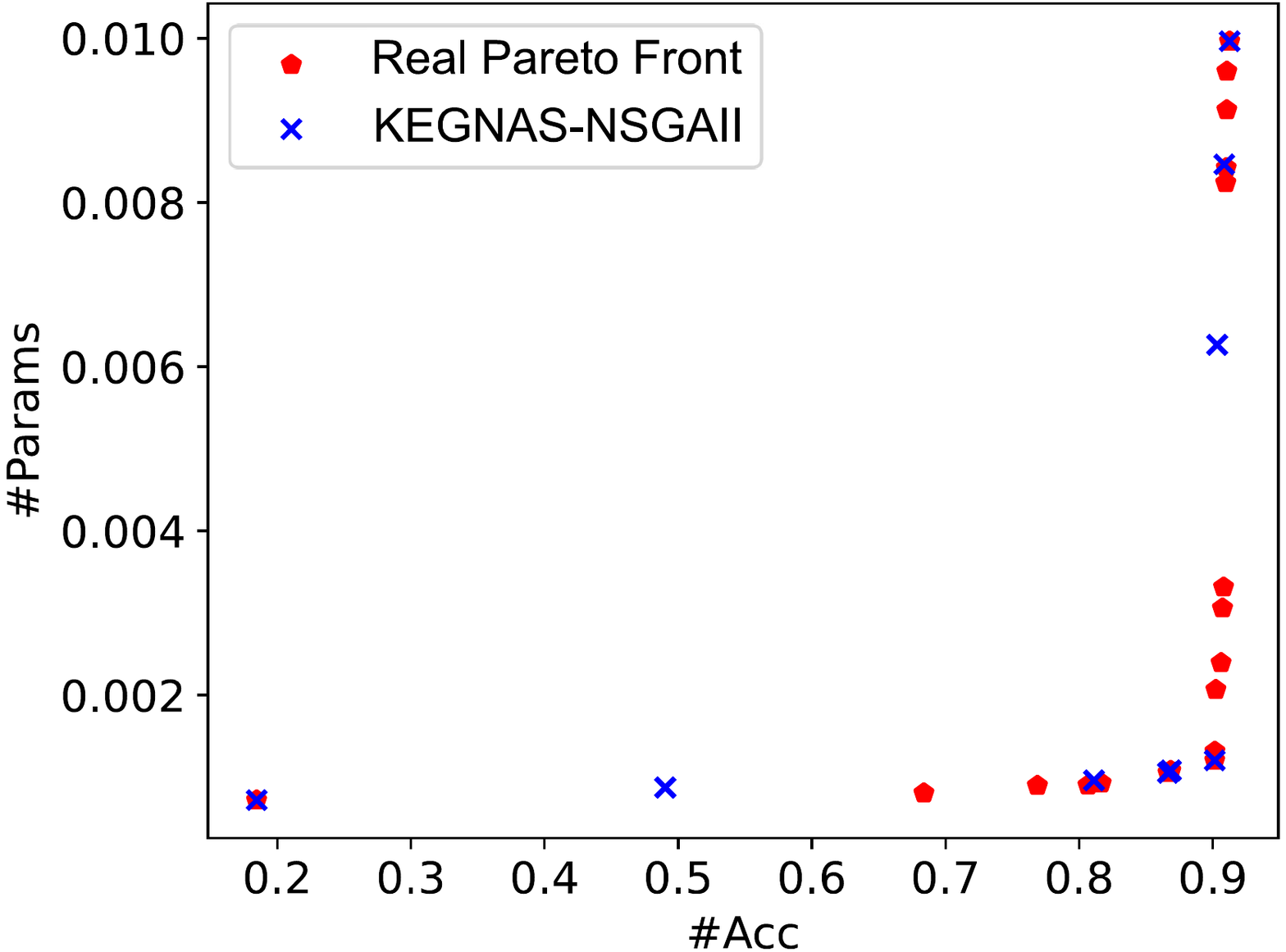}}
	\quad
	\subfloat[Physics]{
		\includegraphics[width=0.36\linewidth]{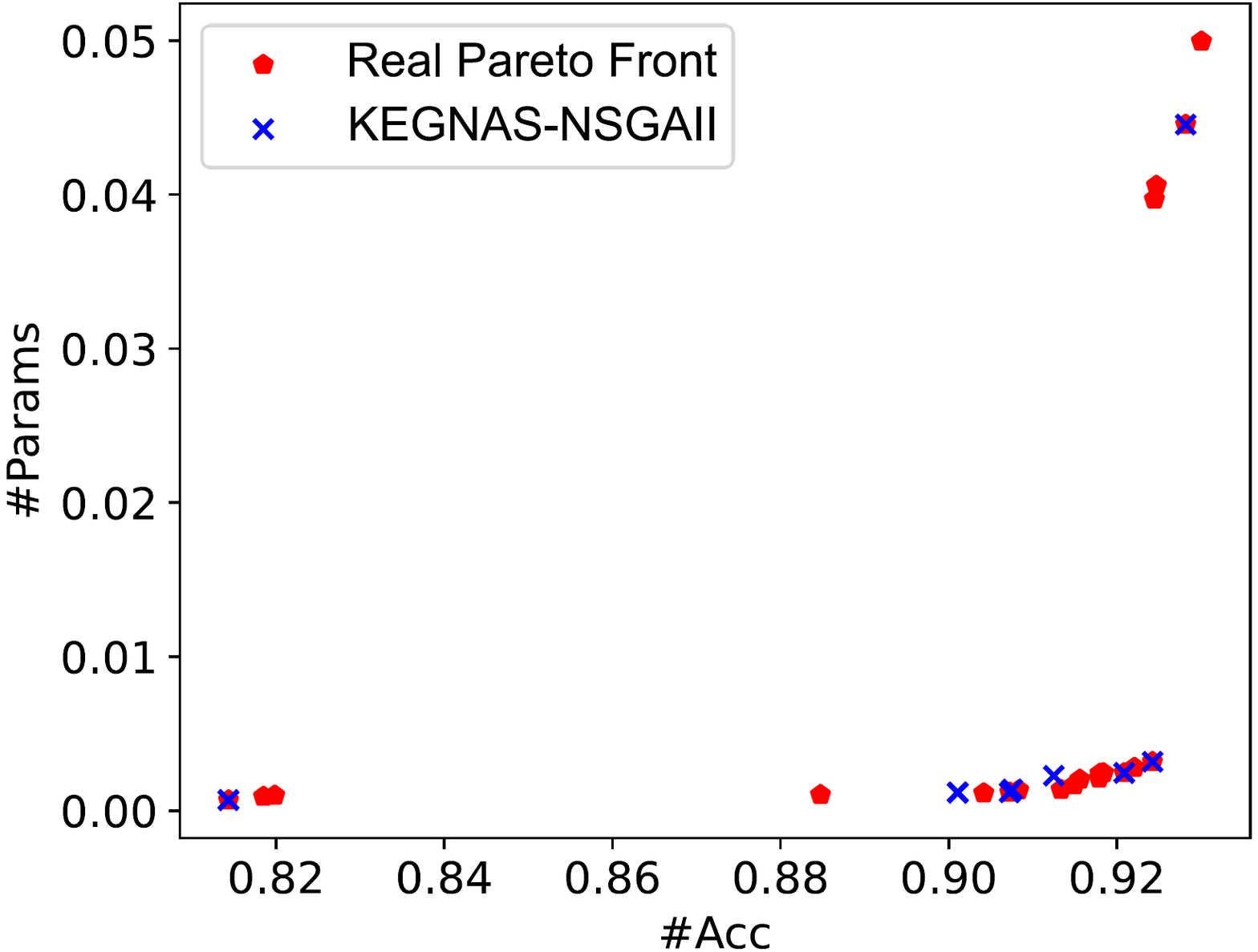}}
	\quad
	\subfloat[Photo]{
		\includegraphics[width=0.36\linewidth]{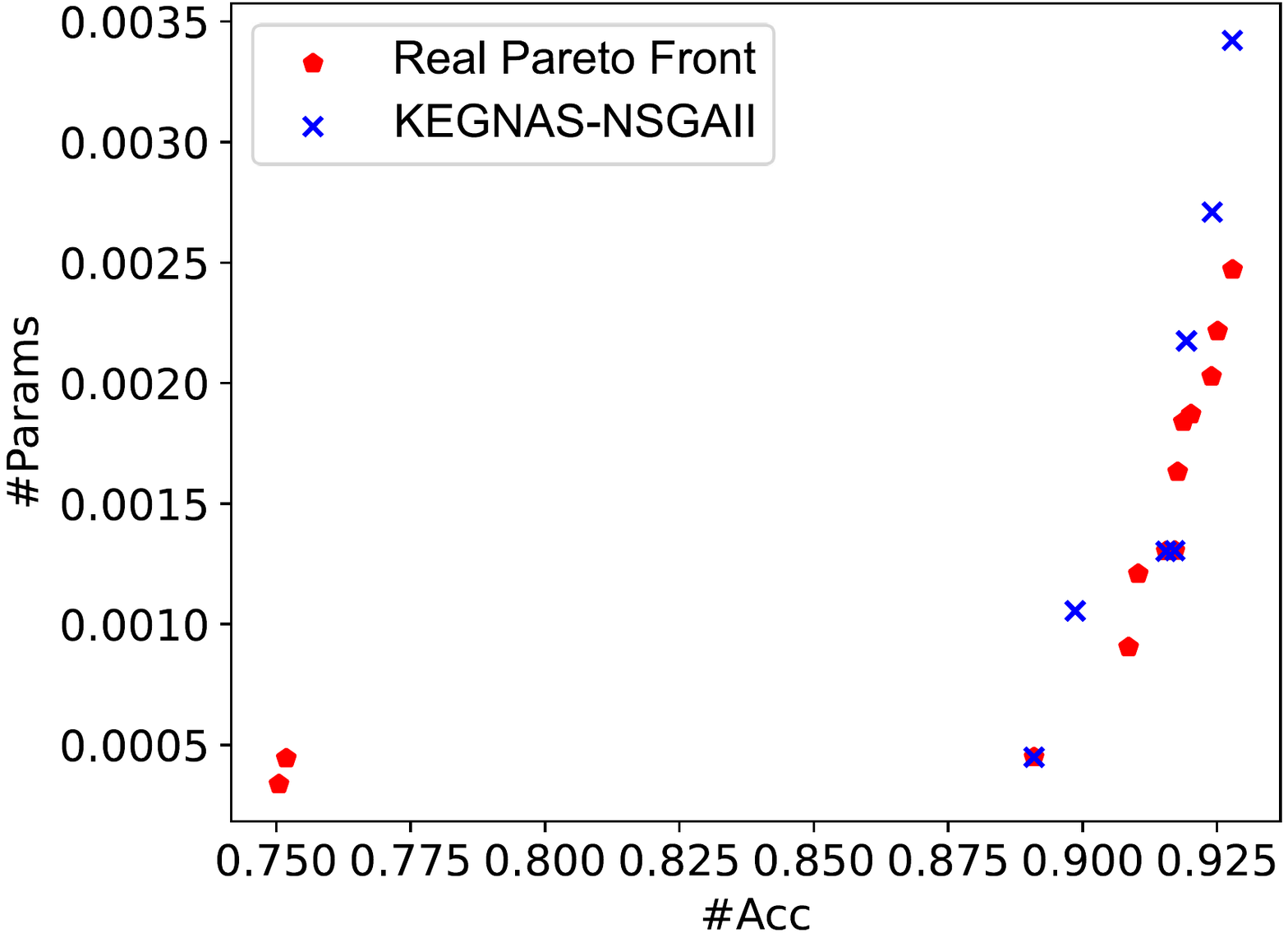}}
    \quad	
    \subfloat[Computers]{
		\includegraphics[width=0.36\linewidth]{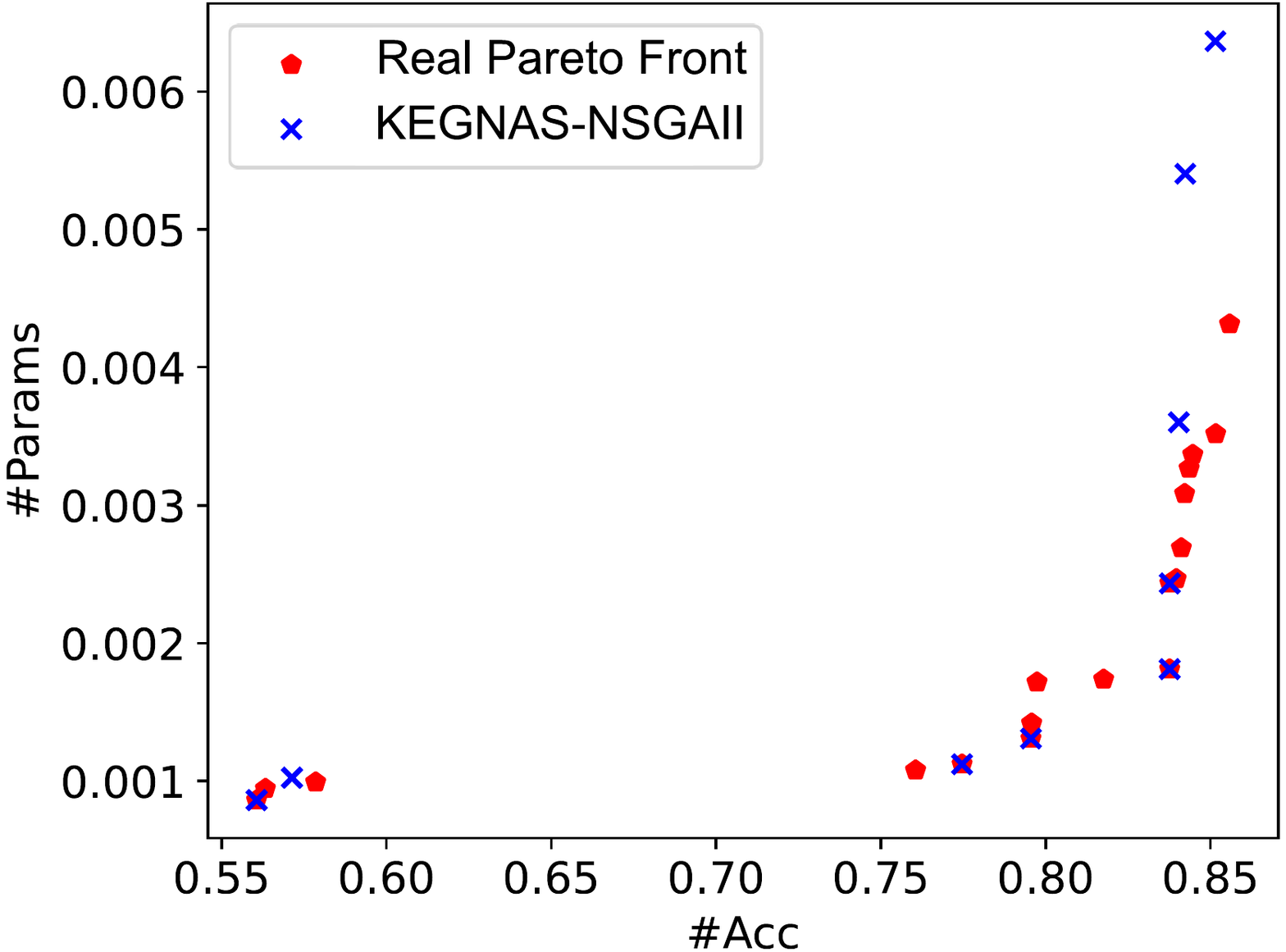}}
    \quad	
    \subfloat[ArXiv]{
		\includegraphics[width=0.36\linewidth]{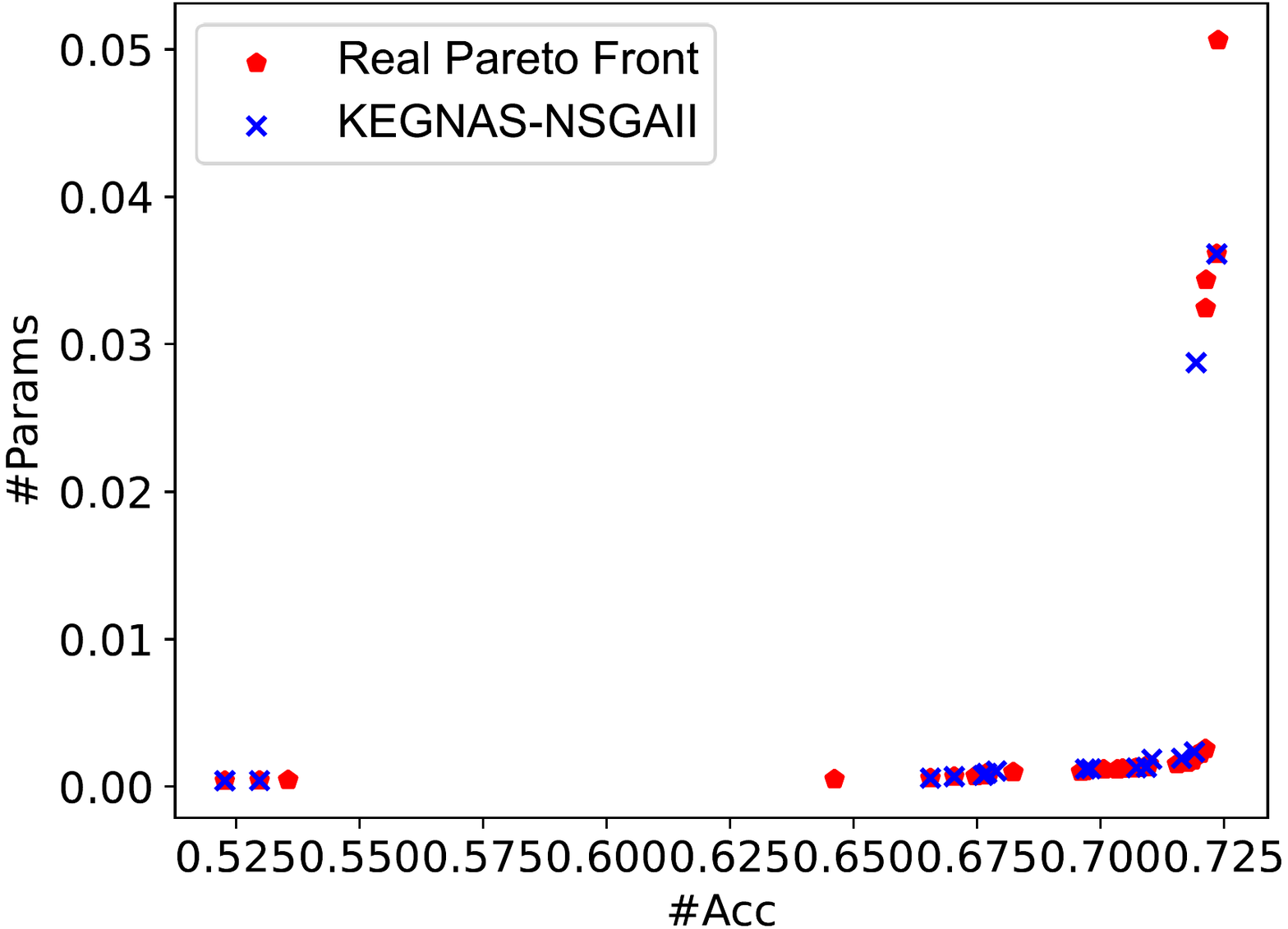}}
    \caption{Comparison of the front (\#Acc and \#Params) obtained by KEGNAS-NSGAII and the true Pareto front on NAS-Bench-Graph.}
	\label{front}
\end{figure*}

Table \ref{Acc_Nas_Bench_Graph} lists the experimental results (average value and standard deviation) of all baselines on NAS-Bench-Graph in terms of \#Acc (\%). For NSGAII and KEGNAS-NSGAII, the architecture with the highest accuracy in the last non-dominated population $\bm{P}_{\bm{A}}^*$ is reported. The best performances are indicated in bold. The average rank (AR) of each baseline is calculated for each dataset.

As shown in Table \ref{Acc_Nas_Bench_Graph}, KEGNAS-NSGAII outperforms the baselines on all datasets, indicating that our proposed method can search for high-performing GNN models in tasks with different similarities. First, KEGNAS-NSGAII completely outperforms NSGAII in terms of \#Acc. This performance gain indicates that the transfer architectures obtained by the knowledge model and DMOGP can be used to improve the search efficiency of NSGAII for higher-quality GNN models. Moreover, compared with the two baselines (GraphNAS and Auto-GNN) specifically tailored for GNAS, KEGNAS-NSGAII still achieves state-of-the-art performance in terms of \#Acc, demonstrating the superiority of our proposal. \textcolor{black}{It can be observed that multiple baselines identify architectures that perform better than the top 5\% architecture, further demonstrating the research necessity of GNAS.}

On many datasets (e.g. the CS dataset), we find that the performance gain of KEGNAS-NSGAII is limited. In addition, KEGNAS-NSGAII can provide decision-makers with an architecture set with different model parameters by optimizing multiple objectives, simultaneously. To investigate the quality of the architecture sets and the reasons for the limited performance gains further, the front (\#Acc and \#Params) obtained by KEGNAS-NSGAII and the true Pareto front are compared in Fig. \ref{front}. The Pareto front in Fig. \ref{front} clearly shows the trade-off between \#Acc and \#Params, which helps decision-makers to understand the different behaviors between the two objectives and how much sacrificing one objective can improve the other objective. The following conclusions can be drawn from Fig. \ref{front}:

\begin{itemize}
    \item On most datasets, KEGNAS-NSGAII identifies the Pareto architecture with the highest accuracy. In addition, the accuracy gap between real Pareto architectures is not obvious. These phenomena result in limited performance gains. To illustrate the superiority of KEGNAS-NSGAII further, we present a performance comparison on real-world datasets in the next section.
    \item On all datasets, the architectures obtained by KEGNAS-NSGAII have multiple overlaps with the Pareto architectures, which shows that our proposal can provide a high-quality architecture set. However, traditional GNAS methods, such as EA and RL, can only obtain a GNN architecture.
    
\end{itemize}

\begin{figure}[ht]
	\centering
	\subfloat[Macro space choices in the real Pareto architecture set]{
		\includegraphics[width=0.4\linewidth]{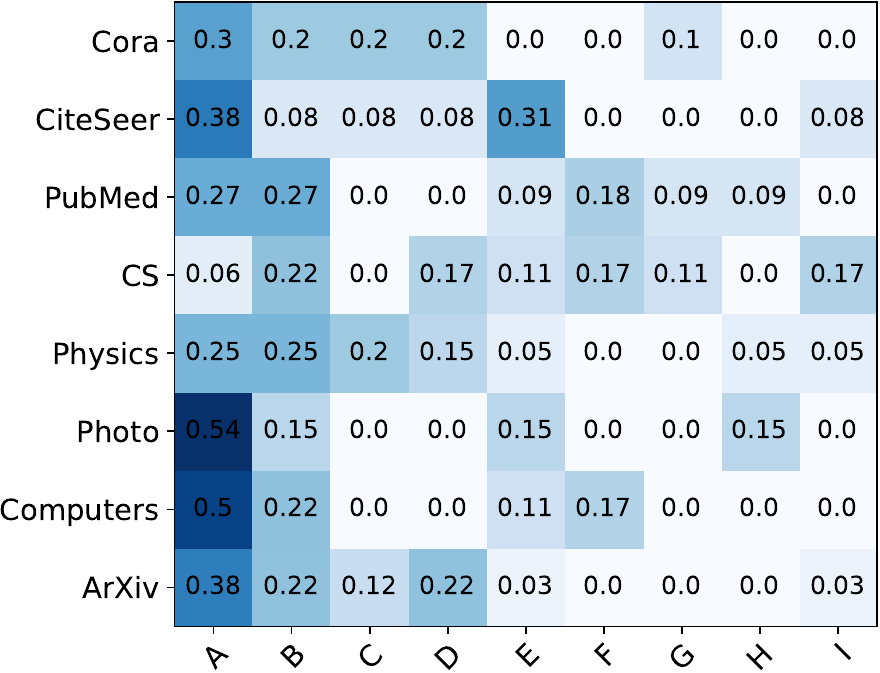}}
	\quad
    \subfloat[Macro space choices in architectures searched by KEGNAS-NSGAII]{
		\includegraphics[width=0.4\linewidth]{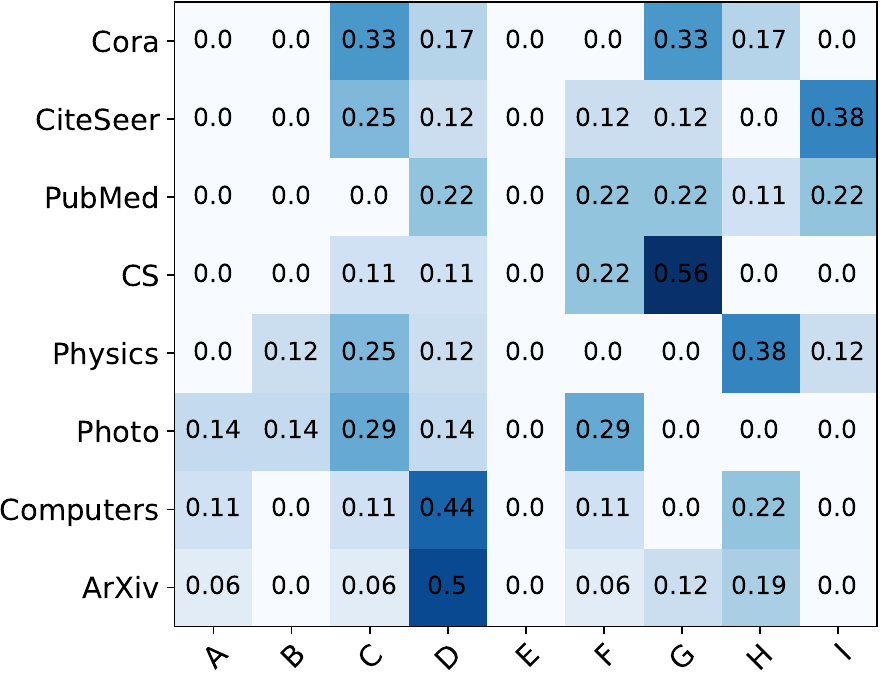}}
	\quad	
    \subfloat[Operation choices in the real Pareto architecture set]{
		\includegraphics[width=0.4\linewidth]{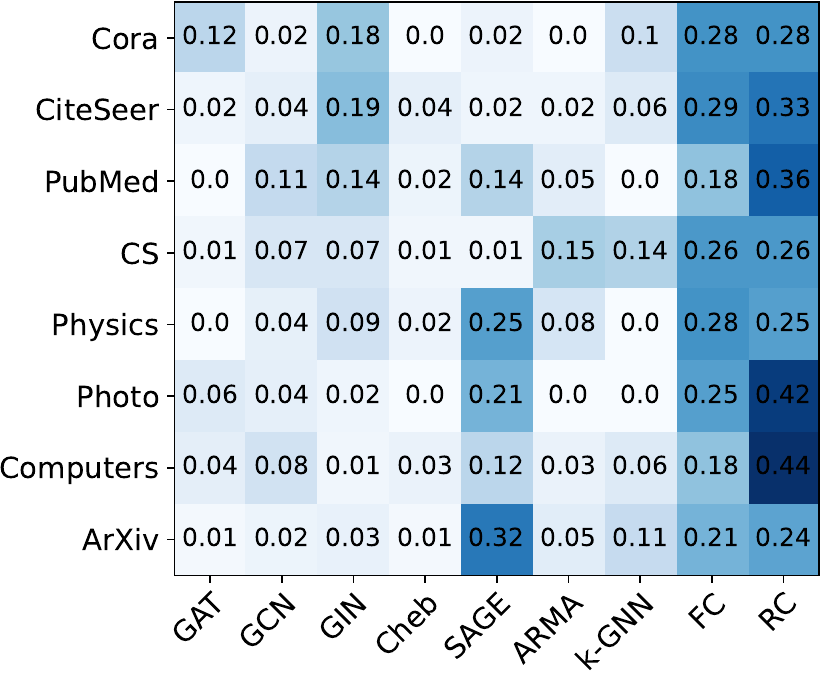}}
	\quad
	\subfloat[Operation choices in architectures searched by KEGNAS-NSGAII]{
		\includegraphics[width=0.4\linewidth]{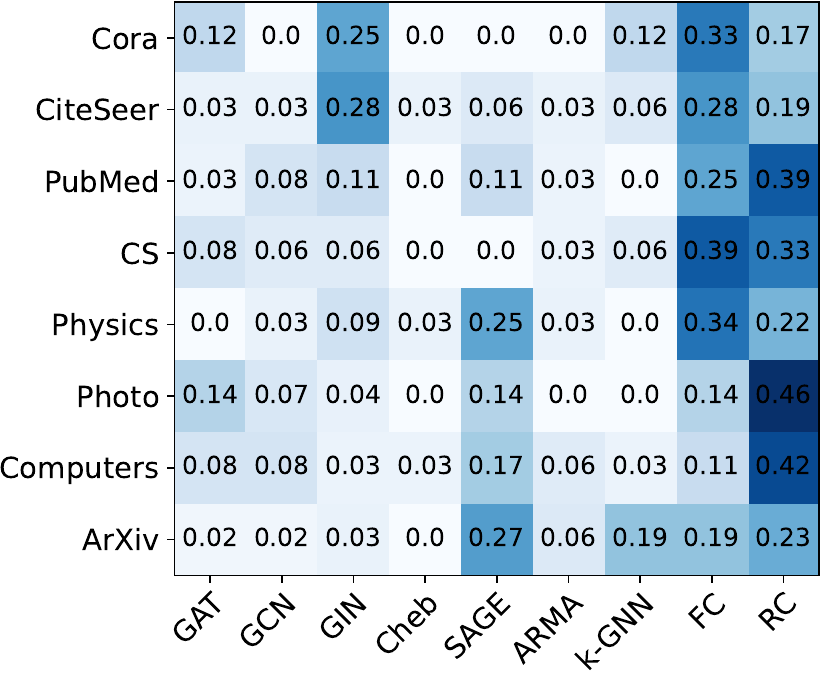}}
	\quad
    \caption{Frequency of the macro space and candidate operations in the real Pareto architecture set and architectures with high accuracy searched by KEGNAS-NSGAII.}
	\label{frequency}
\end{figure}

The search space of NAS-Bench-Graph consists mainly of the macro space and candidate operations. Fig. \ref{frequency} shows the frequency of the macro space and candidate operations in the real Pareto architecture set and architecture set searched by KEGNAS-NSGAII. \textcolor{black}{Significant differences can be observed in the macro space distribution of the searched architectures in different datasets. In particular, small-scale datasets such as Cora, CiteSeer, and PubMed prefer shallow topologies, including F, G, H, and I (for details on these topologies, see \cite{qin2022nasbenchgraph}). This preference aligns with human intuition because simpler models are often sufficient for smaller datasets. Conversely, the other datasets exhibit a more balanced distribution across the macro space, highlighting the importance of GNN architectures tailored for different datasets.} In addition, there are significant differences between the architecture set obtained by KEGNAS-NSGAII and the real Pareto architecture set on small-scale datasets. Therefore, the difference in the macro space may lead to a gap between the front obtained by our proposal and the Pareto front (see Fig. \ref{front} (a)-(c)).

We find that different datasets exhibit similar patterns for the operation distribution. GCN, GIN, FC, and RC are selected in almost all datasets. Although k-GNNs \cite{Morris_Ritzert_Fey_Hamilton_Lenssen_Rattan_Grohe_2019} theoretically perform better in terms of the Weifeiler-Lehman test, they have not been widely selected. Because of the use of the summation aggregation function, k-GNNs may not be suitable for large-scale node-level tasks. \textcolor{black}{On the other hand, the widespread selection of RC in various datasets underscores the crucial role of residuals in enhancing the model performance.}

\subsection{Experimental Studies on Real-world Graph Datasets}

\subsubsection{Experimental Settings}

\begin{table}[htbp]
\scriptsize
\centering
\caption{Details of real-world graph datasets.}\label{datasets_1}
\begin{tabular}{ccccc}
\toprule
Dataset & \#Nodes & \#Edges & \#Features & \#Classes \\
\midrule
\textcolor{black}{Texas} & \textcolor{black}{183} & \textcolor{black}{309} & \textcolor{black}{1,703} & \textcolor{black}{5} \\
Wisconsin & 251 & 466 & 1,703 & 5 \\
\textcolor{black}{Cornell} & \textcolor{black}{183} & \textcolor{black}{298} & \textcolor{black}{1,703} & \textcolor{black}{5} \\
Actor & 7,600 & 30,019 & 932 & 5 \\
Flickr & 89,250 & 899,756 & 500 & 7 \\
\bottomrule
\end{tabular}
\end{table}

\paragraph{Datasets and Tasks} As shown in Table \ref{datasets_1}, we consider \textcolor{black}{five} real-world graph datasets that are widely adopted in the GNAS field: \textcolor{black}{Texas}, Wisconsin, \textcolor{black}{Cornell}, Actor, and Flickr \cite{10.1145/3485447.3512185}. \textcolor{black}{Texas, Wisconsin, and Cornell \footnote{\url{https://github.com/graphdml-uiuc-jlu/geom-gcn}} are the hyperlink networks representing the hyperlink relationship between web pages.} Actor is a social network describing actor relationships, where each node is an actor and each edge represents the co-occurrence of two actors on the same Wikipedia page. Flickr describes common properties between different images, such as the same geographic location. All datasets can be downloaded directly from PyTorch Geometric \textsuperscript{\ref {pyg}}. For \textcolor{black}{Texas, Wisconsin, Cornell, and Actor}, the nodes of each class are randomly split into 48\%, 32\%, and 20\% for $\mathcal{D}_{tra}$, $\mathcal{D}_{val}$, and $\mathcal{D}_{tet}$, respectively \cite{Pei2020Geom}. For Flickr, the nodes of each class are split into 50\%, 25\%, and 25\% for $\mathcal{D}_{tra}$, $\mathcal{D}_{val}$, and $\mathcal{D}_{tet}$, respectively \cite{Zeng2020GraphSAINT}.

For GNAS tasks on a real-world graph dataset, the GNAS task on all datasets in NAS-Bench-Graph can be used as the source tasks. For example, for the GNAS task on the Actor dataset, GNAS tasks on the eight datasets are regarded as eight source tasks, thus constituting the prior database $\mathcal{M}$. \textcolor{black}{The knowledge model and DMOGP only need to be trained on $\mathcal{M}$ once. These trained models can be utilized directly by users.
For each GNAS task on the real-world graph dataset, the knowledge model and DMOGP can directly generate and evaluate transfer architectures in only a few seconds.} When constructing the prior database, similar to the experimental settings in the previous section, the number of non-dominated fronts $N_s$ for each source task is set to 10.

\paragraph{Baselines} We provide \textcolor{black}{six} different types of state-of-the-art baselines: Random (a random search), F2GNN (a differentiable method) \cite{10.1145/3485447.3512185}, SANE (a differentiable method) \cite{9458743} SNAG (an RL-based method) \cite{zhao2020simplifying}, \textcolor{black}{Genetic-GNN (an EA-based method) \cite{SHI2022108752}, and CTFGNAS (an EA-based method) \cite{LIU2023110485}}. Both SANE and F2GNN are advanced differentiable methods. \textcolor{black}{They focus on studying the impact of aggregation operations and topology on GNN performance. SNAG is an advanced RL-based method for searching for aggregation operations and connections.} \textcolor{black}{Genetic-GNN is a popular evolutionary GNAS method, which collaboratively optimizes the architecture and hyperparameters. CTFGNAS is a state-of-the-art surrogate-assisted evolutionary GNAS algorithm used for exploring layer components, topological connections, and fusion strategies.}

\paragraph{Implementation Details} The experimental environment is consistent with that described in the previous section. For F2GNN \footnote{\url{https://github.com/LARS-research/F2GNN}}, SANE \footnote{\url{https://github.com/LARS-research/SANE}}, SNAG \footnote{\url{https://github.com/LARS-research/SNAG}}, \textcolor{black}{Genetic-GNN\footnote{\url{https://github.com/codeshareabc/Genetic-GNN}}, and CTFGNAS\footnote{\url{https://github.com/chnyliu/CTFGNAS}}}, we adopt the public codes provided by the original authors. In line with the existing method \cite{10.1145/3485447.3512185,9458743}, the number of independent runs is set to 10 for all baselines. For a fair comparison,  the number of backbones in the baselines is set to 4. 

For Random, 100 architectures are randomly sampled from the search space of NAS-Bench-Graph and trained from scratch. The architecture with the highest accuracy is selected.

For \textcolor{black}{F2GNN and SANE}, in the search phase, we use an Adam optimizer with a learning rate of 0.001 and a batch size of 256 for 400 epochs.

For \textcolor{black}{SNAG}, the search epochs are set to 400. In each epoch, we sample and evaluate 10 architectures to update the controller. Finally, the controller is employed to generate the final architecture.

\textcolor{black}{For Genetic-GNN and CTFGNAS, all hyperparameters are set to the recommended values in the original paper.} 

For KEGNAS-NSGAII, the generation and evaluation of transfer architectures require only a small computational cost. However, during the search process of NSGAII (Line 15 in Algorithm \ref{alg2}), evaluating the architectures in the population on real-world graph datasets is expensive. Therefore, we modify KEGNAS-NSGAII to reduce the search cost. Specifically, we first train all transfer architectures on real-world graph datasets. Subsequently, Gaussian models are constructed using the transfer architectures and their real performance metrics. Finally, Gaussian models are employed as surrogate models to replace the evaluation process in NSGAII to save computational resources. We set the number of epochs to 100 when training each candidate transfer architecture. The number of candidate transfer architectures, $N_c$, is set to 500. The population size, $N_p$, is set to 20. The maximum number of generations, $G$, is set to 100.
Finally, we select the architecture with the highest accuracy from the non-dominated architectures $\bm{P}_{\bm{A}}^*$ as the final result.

We retrain the searched architectures using each baseline from scratch for 400 epochs. Consistent with existing methods \cite{10.1145/3485447.3512185}, we adopt Hyperopt \footnote{\url{https://github.com/hyperopt/hyperopt}} to optimize the hyperparameters of the architecture, \textcolor{black}{the search space of which is outlined in Table \ref{hyperparameter}.} The number of iterations for the hyperparameter optimization is set to 50. The optimal hyperparameters of the validation set are adopted to train the searched architecture.

\begin{table}[htbp]
\scriptsize
\centering
\caption{Hyperparameters used in retraining.}\label{hyperparameter}
\begin{tabular}{cc}
\toprule
Hyperparameter & Operations \\
\midrule
\#Embedding Size & $[16,32,64,128,256,512]$ \\
\#Learning Rate &  $[0.01, 0.001]$  \\
\#Optimizer & [Adam, AdaGrad] \\
\bottomrule
\end{tabular}
\end{table}

\subsubsection{Experimental Results}

\begin{table}[htbp]
\begin{center}
\tiny
\color{black}
\caption{Experimental results of all baselines on real-world graph datasets in terms of \#Acc (\%). ``OOM" means out of memory.}
\label{Acc_RWGD}
\begin{tabular}{cccccc|c}
\toprule
Baseline & Texas & Wisconsin & Cornell & Actor & Flickr & AR \\
\midrule
        Random & 71.24(8.17) & 84.71(3.37) & 43.52(3.54) & 36.34(1.21) & 53.06(0.11) & 4.4 \\
        F2GNN & 82.97(5.41) & 84.12(5.71) & 76.27(4.95) & 25.24(1.13) & 47.23(0.21) & 4.6 \\
        SANE & 68.38(5.57) & 85.10(3.30) & 55.78(5.63) & 36.64(1.30) & 51.98(0.09) & 4.2 \\
        SNAG & 77.79(5.21) & 43.92(4.34) & 70.37(4.90) & 26.92(0.98) & 47.21(0.10) & 5.6 \\
        Genetic-GNN & 86.65(4.53) & 62.51(1.74) & 57.40(2.12) & 32.10(2.14) & OOM & 5.2 \\
        CTFGNAS & 87.57(4.71) & 78.83(4.00) & 78.42(2.96) & \textbf{38.12(0.43)} & 51.28(0.38) & 2.8 \\
        KEGNAS-NSGAII & \textbf{90.81(1.32)} & \textbf{90.98(2.00)} & \textbf{82.70(2.48)} & 37.84(0.69) & \textbf{53.25(0.65)} & 1.2 \\
\bottomrule
\end{tabular}
\end{center}
\end{table}

Table \ref{Acc_RWGD} lists the experimental results (average value and standard deviation) of all baselines on the real-world graph datasets in terms of \#Acc (\%). For KEGNAS-NSGAII, the architecture with the highest accuracy in the last non-dominated population $\bm{P}_{\bm{A}}^*$ is reported. \textcolor{black}{Owing to the utilization of information in NAS-Bench-Graph, KEGNAS-NSGAII outperforms all comparative baselines overall, including the state-of-the-art differentiable, RL-based, and single-objective EA-based methods. Notably, KEGNAS-NSGAII achieves an average improvement of 4.27\% over CTFGNAS, the best baseline, and an impressive improvement of 11.54\% over SANE, the leading differentiable method.}
This phenomenon indicates the capability of the MOEA in designing GNNs. By utilizing the transfer architectures obtained by the knowledge model and DMOGP, our proposed method achieves considerable performance gains, \textcolor{black}{demonstrating the importance of prior knowledge in improving model performance.} Furthermore, KEGNAS-NSGAII outperforms Random, illustrating the usefulness of the improved search algorithm for designing the GNN architecture.

\begin{table}[ht]
\begin{center}
\tiny
\color{black}
\caption{Comparison of the search cost (GPU seconds) of KEGNAS-NSGAII and baselines.}\label{search-time}
\begin{tabular}{cccccc}
\toprule
Dataset & Baselines & AR & Pre-search & During-search & Total \\
\midrule
 \multirow{5}{*}{Texas} & Random & 6 & - & 6.2K & 6.2K \\
 & F2GNN & 4 & - & 0.4K & \textbf{0.4K} \\
 & SANE & 7 & - & 0.4K & \textbf{0.4K} \\
 & SNAG & 5 & - & 5.6K & 5.6K \\
 & Genetic-GNN & 3 & - & 64.9K & 64.9K \\
 & CTFGNAS & 2 & - & 10.7K & 10.7K \\
 & KEGNAS-NSGAII & \textbf{1} & \textbf{2.0}+1.0K & 19.2 & 1.0K \\
 \midrule
 \multirow{5}{*}{Wisconsin} & Random & 3 & - & 133.7K & 133.7K \\
 & F2GNN & 4 & - & 2.2K & 2.2K \\
 & SANE & 2 & - & 1.6K & \textbf{1.6K} \\
 & SNAG & 7 & - & 2,806.4K & 2,806.4K \\
 & Genetic-GNN & 6 & - & 1,549.4K & 1,549.4K \\
 & CTFGNAS & 5 & - & 17.8K & 17.8K \\
 & KEGNAS-NSGAII & \textbf{1} & \textbf{0.6}+12.0K & 23.6 & 12.0K \\
 \midrule
 \multirow{5}{*}{Cornell} & Random & 7 & - & 5.2K & 5.2K \\
 & F2GNN & 3 & - & 0.4K & \textbf{0.4K} \\
 & SANE & 6 & - & 0.4K & \textbf{0.4K} \\
 & SNAG & 4 & - & 6.1K & 6.1K \\
 & Genetic-GNN & 5 & - & 54.K & 54.K \\
 & CTFGNAS & 2 & - & 9.2K & 9.2K \\
 & KEGNAS-NSGAII & \textbf{1} & \textbf{2.1}+1.0K & 21.8 & 1.0K \\
\midrule
\multirow{5}{*}{Actor} & Random & 4 & - & 131.6K & 131.6K \\
 & F2GNN & 7 & - & 3.6K & 3.6K \\
 & SANE & 3 & - & 1.0K & \textbf{1.0K} \\
 & SNAG & 6 & - & 1,842.4K & 1,842.4K \\
 & Genetic-GNN & 5 & - & 1,527.2K & 1,527.2K \\
 & CTFGNAS & \textbf{1} & - & 28.6K & 28.6K \\
 & KEGNAS-NSGAII & 2 & \textbf{1.2}+26.2K & 43.4 & 26.2K \\
 \midrule
 \multirow{5}{*}{Flickr} & Random & 2 & - & 254.6K & 254.6K \\
 & F2GNN & 5 & - & 4.2K & 4.2K \\
 & SANE & 3 & - & 4.0K & \textbf{4.0K} \\
 & SNAG & 6 & - & 6,950.8K & 6,950.8K \\
 & Genetic-GNN & 7 & - & - & OOM \\
 & CTFGNAS & 4 & - & 45.0k & 45.0k \\
 & KEGNAS-NSGAII & \textbf{1} & \textbf{2.1}+32.0K & 31.8 & 32.0K \\
\bottomrule
\end{tabular}
\end{center}
\end{table}

The total search cost incurred by the baseline can be divided into two stages: (1) pre-search, which is the search cost incurred prior to the search, and (2) during-search, which is the cost of executing the search algorithm. Table \ref{search-time} lists the average search times of the baselines for the different datasets. ``–" indicates not applicable. 
For KEGNAS-NSGAII, the pre-search consists of two parts: 1) the generation and evaluation of the transfer architectures, and 2) the construction of the surrogate model used in NSGAII. Table\ref{search-time} shows that the cost of pre-search is mainly consumed by the latter. Notably, generating and evaluating candidate transfer architectures requires only a few GPU seconds in total, which is negligible compared with the search cost of NSGAII.

Compared with the Random and RL-based (SNAG) baselines, KEGNAS-NSGAII achieves better overall performance in less time by utilizing transfer architectures rich in prior knowledge, demonstrating the high efficiency of our proposal. The search cost of the differentiable methods (F2GNN and SANE) is lower than that of KEGNAS-NSGAII. \textcolor{black}{However, Table \ref{Acc_RWGD} shows that the proposed method performs better in terms of \#Acc and robustness.} \textcolor{black}{In Genetic-GNN, the real fitness evaluation involves model training over multiple epochs, rendering its search cost significantly higher than that of KEGNAS-NSGAII. Furthermore, compared with the current state-of-the-art surrogate-assisted EA (CTFGNAS), KEGNAS-NSGAII identifies architectures with superior performance at a lower computational cost, further highlighting the effectiveness of prior knowledge assistance.} 
It should be emphasized that KEGNAS is more flexible. KEGNAS can easily be embedded in MOEAs and does not require differentiable approximations or assumptions.

\subsection{Effectiveness of Prior Knowledge in KEGNAS}

\begin{table}[htbp]
\begin{center}
\caption{Average HV values of the architecture sets obtained by all baselines on NAS-Bench-Graph over 20 runs.}
\label{HV_Nas_Bench_Graph}
\scriptsize
\begin{tabular}{cccc}
\toprule
Baseline & NSGAII & RKEGNAS-NSGAII & KEGNAS-NSGAII \\
\midrule
        Cora & 0.26(0.10)$-$ & 0.25(0.10)$-$ & \textbf{0.33(0.11)} \\
        CiteSeer & 0.24(0.08)$-$ & 0.22(0.08)$-$ & \textbf{0.31(0.08)} \\
        PubMed & 0.12(0.01)$\approx$ & 0.13(0.01)$\approx$ & \textbf{0.14(0.02)} \\
        CS & 0.71(0.16)$-$ & 0.63(0.22)$-$ & \textbf{0.78(0.12)} \\
        Physics & 0.19(0.03)$-$ & 0.18(0.04)$-$ & \textbf{0.43(0.10)} \\
        Photo & 0.21(0.05)$\approx$ & 0.20(0.06)$-$ & \textbf{0.25(0.05)} \\
        Computers & 0.31(0.06)$-$ & 0.32(0.04)$-$ & \textbf{0.37(0.03)} \\
        ArXiv & 0.28(0.07)$-$ & 0.30(0.07)$\approx$ & \textbf{0.34(0.07)} \\
\midrule
$+/\approx/-$ & $0/2/6$ & $0/2/6$ & $-$ \\
\bottomrule
\end{tabular}
\end{center}
\end{table}

We perform a series of experimental studies on the NAS-Bench-Graph to explore whether prior knowledge can improve the search efficiency. In addition to KEGNAS-NSGAII, two baselines, NSGAII and RKEGNAS-NSGAII, are constructed to verify the effectiveness of the knowledge model and DMOGP, respectively. For NSGAII, the initial population is randomly generated, which means that this is a zero-knowledge state. For RKEGNAS-NSGAII, the transfer architectures are randomly selected from the candidate transfer architectures without using DMOGP. HV, a popular metric\cite{1688440}, is employed to evaluate the architecture set in multi-objective machine learning. In general, architecture with a larger HV is considered to be of better quality. Table \ref{HV_Nas_Bench_Graph} presents the average HV values of the architecture sets obtained by all baselines on the NAS-Bench-Graph over 20 runs. According to the Wilcoxon rank-sum test with a 95\% confidence level, \textcolor{black}{the symbols ``$+/\approx/-$" indicate that the average HV values of the corresponding baselines are significantly better than/similar to/worse than that of KEGNAS-NSGAII.}

According to the comparison results for NAS-Bench-Graph shown in Table \ref{HV_Nas_Bench_Graph}, KEGNAS-NSGAII exhibits excellent performance in all cases in terms of the average HV value. Compared with NSGAII, our proposed method performs better in 6 and ties 2 out of 8 cases. In NSGAII, the initial population contains no prior knowledge. In KEGNAS-NSGAII, the transfer architectures generated by the knowledge model are employed to warm-start NSGAII. Given that NSGA-II and KEGNAS-NSGA-II have the same fundamental optimization solver, yet differ primarily in transfer architectures designed to incorporate knowledge, the effectiveness of the knowledge model in KEGNAS-NSGA-II is confirmed.
This phenomenon also shows that transfer architectures carrying prior knowledge can significantly improve the search performance. It is worth noting that the knowledge model and DMOGP do not need to be retrained for each new dataset. \textcolor{black}{Therefore, only a few GPU seconds are required to produce and evaluate candidate transfer architectures, which is trivial relative to the subsequent search time.}

In addition, KEGNAS-NSGAII exceeds or matches RKEGNAS-NSGAII in all cases. In KEGNAS, DMOGP is used only to evaluate the candidate transfer architectures for selecting transfer architectures. Because RKEGNAS-NSGAII and KEGNAS-NSGAII have the same components apart from the transfer architecture selection, the performance gain demonstrates the effectiveness of the DMOGP.

\subsection{Key Parameter Analysis}

\begin{figure}[htbp]
\centering
\includegraphics[width=0.8\textwidth]{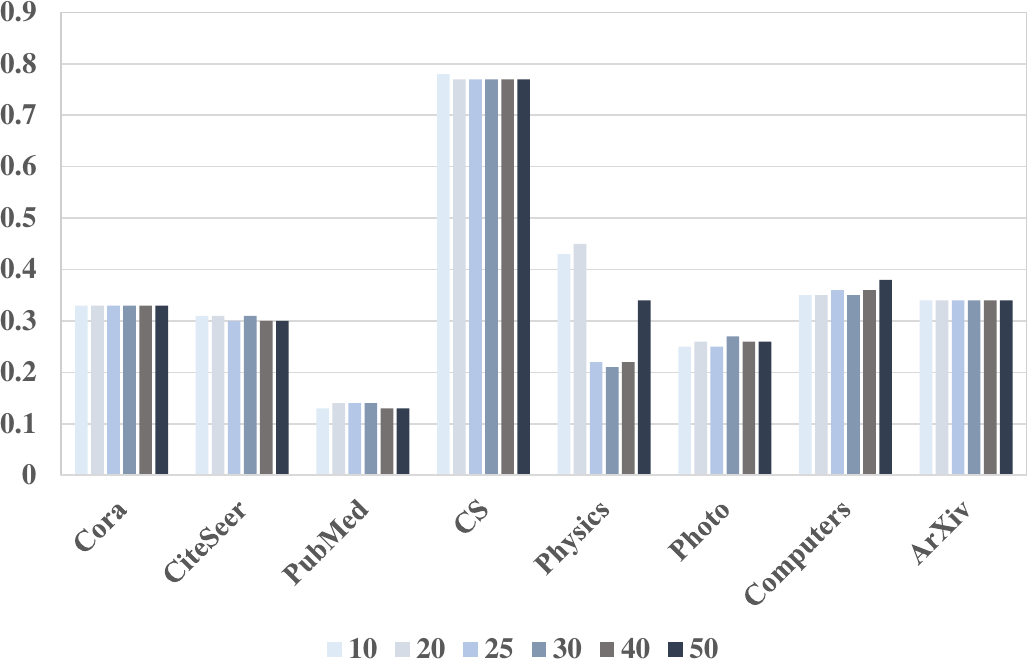}
\caption{Average HV values with various configurations ($[10,20,25,30,40,50]$) of $N_p$ over 20 runs.} \label{pop_size}
\end{figure}

In this section, we describe an experiment using NAS-Bench-Graph to analyze the population size $N_p$ of KEGNAS-NSGAII. Fig. \ref{pop_size} shows the average HV values with various configurations ($[10,20,25,30,40,50]$) of $N_p$ over 20 runs. The maximum number of evaluations is 500. In the figures for the analysis of parameter $N_p$, the x-axis represents the datasets and the y-axis represents the average HV value. Given the maximum number of evaluations, the different configurations of $N_p$ do not significantly affect the performance of KEGNAS-NSGAII as shown in Fig. \ref{pop_size}. It is recommended to set $N_p$ to a small value to obtain better performance on all datasets.

\section{Conclusions and Future Work}
\label{sec6}
This study aimed to establish a new ETO paradigm, KEGNAS, which improves GNAS performance by transferring valuable prior knowledge. In particular, it is demonstrated that GNAS problems can benefit from knowledge bases such as NAS-Bench-Graph. After formalizing the problem, three components are proposed: the knowledge model, DMOGP, and the warm-start MOEA. By leveraging the prior knowledge in NAS-Bench-Graph, the knowledge model and DMOGP can quickly generate and evaluate candidate transfer architectures for unseen GNAS tasks. These transfer architectures can significantly improve the search capability of the MOEA. A series of experimental results on NAS-Bench-Graph and real-world graph datasets reveal the remarkable efficacy of KEGNAS. In addition, we empirically verify the effectiveness of prior knowledge utilization through ablation studies.

\textcolor{black}{Although KEGNAS has shown encouraging preliminary outcomes, it is essential to acknowledge that the applicability of knowledge transfer may vary across different types of graph tasks, such as link prediction and graph classification. Our knowledge model and DMOGP may encounter two key challenges: the acquisition of a prior knowledge base for constructing training data tailored to these tasks and the fine-tuning of the model architecture to match the task-specific search space. Nevertheless, KEGNAS offers a novel perspective on transfer optimization within the GNAS framework, yielding significant enhancements in search efficiency and performance when applied to real-world graph datasets.}

\textcolor{black}{Several key research directions should be explored to advance the KEGNAS framework further.} First, other prior knowledge in GNAS problems, such as the fitness landscape of performance metrics, requires further investigation. Second, this study mainly focuses on the isomorphic search space, that is, the source and target tasks are in the same NAS-Bench-Graph's search space. Knowledge transfer in heterogeneous search spaces remains an open problem that could enhance the applicability of the KEGNAS. \textcolor{black}{Third, the complexity analysis of KEGNAS, particularly its knowledge model and DMOGP, will be a central focus of our future research endeavors. \textcolor{black}{Fourth, we will investigate the compatibility of KEGNAS with various MOEAs.} Finally, establishing rich GNAS benchmarks is considered a valuable research direction because it can provide diverse training data for the knowledge model and DMOGP.}

\section*{Acknowledgments}
\textcolor{black}{This work was supported in part by the Key Scientific Technological Innovation Research Project of the Ministry of Education, the Joint Funds of the National Natural Science Foundation of China (U22B2054), the National Natural Science Foundation of China (62076192, 62276201, and 62276199), the 111 Project, the Program for Cheung Kong Scholars and Innovative Research Team in University (IRT 15R53), and the Science and Technology Innovation Project from the Chinese Ministry of Education, National Key Laboratory of Human-Machine Hybrid Augmented Intelligence, Xi'an Jiaotong University (No. HMHAI-202404).}


 \bibliographystyle{elsarticle-num} 
 \bibliography{my}

\begin{thebibliography}{10}
\expandafter\ifx\csname url\endcsname\relax
  \def\url#1{\texttt{#1}}\fi
\expandafter\ifx\csname urlprefix\endcsname\relax\def\urlprefix{URL }\fi
\expandafter\ifx\csname href\endcsname\relax
  \def\href#1#2{#2} \def\path#1{#1}\fi

\bibitem{10026151}
J.~Chen, B.~Li, K.~He, \href{https://www.sciencedirect.com/science/article/pii/S0950705124004957}{Neighborhood convolutional graph neural network}, Knowledge-Based Systems 295 (2024) 111861.
\newblock \href {https://doi.org/https://doi.org/10.1016/j.knosys.2024.111861} {\path{doi:https://doi.org/10.1016/j.knosys.2024.111861}}.
\newline\urlprefix\url{https://www.sciencedirect.com/science/article/pii/S0950705124004957}

\bibitem{kipf2017semisupervised}
T.~N. Kipf, M.~Welling, Semi-supervised classification with graph convolutional networks, in: International Conference on Learning Representations, 2017.

\bibitem{hamilton2017inductive}
W.~Hamilton, Z.~Ying, J.~Leskovec, Inductive representation learning on large graphs, in: Advances in Neural Information Processing Systems, Vol.~30, Curran Associates, Inc., 2017.

\bibitem{veličković2018graph}
P.~Veličković, G.~Cucurull, A.~Casanova, A.~Romero, P.~Liò, Y.~Bengio, Graph attention networks, in: International Conference on Learning Representations, 2018.

\bibitem{xu2018how}
K.~Xu, W.~Hu, J.~Leskovec, S.~Jegelka, How powerful are graph neural networks?, in: International Conference on Learning Representations, 2019.

\bibitem{9046288}
Z.~Wu, S.~Pan, F.~Chen, G.~Long, C.~Zhang, P.~S. Yu, A comprehensive survey on graph neural networks, IEEE Transactions on Neural Networks and Learning Systems 32~(1) (2021) 4--24.
\newblock \href {https://doi.org/10.1109/TNNLS.2020.2978386} {\path{doi:10.1109/TNNLS.2020.2978386}}.

\bibitem{wang2021automated}
X.~Wang, W.~Zhu, Automated machine learning on graph, in: Proceedings of the 27th ACM SIGKDD Conference on Knowledge Discovery \& Data Mining, ACM, 2021, pp. 4082--4083.

\bibitem{9782531}
Y.~Gao, P.~Zhang, H.~Yang, C.~Zhou, Y.~Hu, Z.~Tian, Z.~Li, J.~Zhou, Graphnas++: Distributed architecture search for graph neural networks, IEEE Transactions on Knowledge and Data Engineering 35~(7) (2023) 6973--6987.
\newblock \href {https://doi.org/10.1109/TKDE.2022.3178153} {\path{doi:10.1109/TKDE.2022.3178153}}.

\bibitem{9714826}
J.~Chen, J.~Gao, Y.~Chen, B.~M. Oloulade, T.~Lyu, Z.~Li, Auto-gnas: A parallel graph neural architecture search framework, IEEE Transactions on Parallel and Distributed Systems 33~(11) (2022) 3117--3128.
\newblock \href {https://doi.org/10.1109/TPDS.2022.3151895} {\path{doi:10.1109/TPDS.2022.3151895}}.

\bibitem{10.1145/3485447.3512185}
L.~Wei, H.~Zhao, Z.~He, Designing the topology of graph neural networks: A novel feature fusion perspective, in: Proceedings of the ACM Web Conference 2022, WWW '22, ACM, 2022, p. 1381–1391.
\newblock \href {https://doi.org/10.1145/3485447.3512185} {\path{doi:10.1145/3485447.3512185}}.

\bibitem{GM2024112145}
B.~G.M., G.~Pillai, \href{https://www.sciencedirect.com/science/article/pii/S0950705124007792}{Sequential node search for faster neural architecture search}, Knowledge-Based Systems 300 (2024) 112145.
\newblock \href {https://doi.org/https://doi.org/10.1016/j.knosys.2024.112145} {\path{doi:https://doi.org/10.1016/j.knosys.2024.112145}}.
\newline\urlprefix\url{https://www.sciencedirect.com/science/article/pii/S0950705124007792}

\bibitem{AN2023110341}
Y.~An, C.~Zhang, X.~Zheng, \href{https://www.sciencedirect.com/science/article/pii/S0950705123000916}{Knowledge reconstruction assisted evolutionary algorithm for neural network architecture search}, Knowledge-Based Systems 264 (2023) 110341.
\newblock \href {https://doi.org/https://doi.org/10.1016/j.knosys.2023.110341} {\path{doi:https://doi.org/10.1016/j.knosys.2023.110341}}.
\newline\urlprefix\url{https://www.sciencedirect.com/science/article/pii/S0950705123000916}

\bibitem{9458743}
H.~Zhao, Q.~Yao, W.~Tu, Search to aggregate neighborhood for graph neural network, in: IEEE International Conference on Data Engineering, IEEE, 2021, pp. 552--563.
\newblock \href {https://doi.org/10.1109/ICDE51399.2021.00054} {\path{doi:10.1109/ICDE51399.2021.00054}}.

\bibitem{shala2023transfer}
G.~Shala, T.~Elsken, F.~Hutter, J.~Grabocka, Transfer {NAS} with meta-learned bayesian surrogates, in: International Conference on Learning Representations, 2023.

\bibitem{SHI2022108752}
M.~Shi, Y.~Tang, X.~Zhu, Y.~Huang, D.~Wilson, Y.~Zhuang, J.~Liu, Genetic-gnn: Evolutionary architecture search for graph neural networks, Knowledge-Based Systems 247 (2022) 108752.
\newblock \href {https://doi.org/https://doi.org/10.1016/j.knosys.2022.108752} {\path{doi:https://doi.org/10.1016/j.knosys.2022.108752}}.

\bibitem{10065594}
C.~Wang, L.~Jiao, J.~Zhao, L.~Li, X.~Liu, F.~Liu, S.~Yang, Bi-level multiobjective evolutionary learning: A case study on multitask graph neural topology search, IEEE Transactions on Evolutionary Computation 28~(1) (2024) 208--222.
\newblock \href {https://doi.org/10.1109/TEVC.2023.3255263} {\path{doi:10.1109/TEVC.2023.3255263}}.

\bibitem{10040227}
Y.~Gao, P.~Zhang, C.~Zhou, H.~Yang, Z.~Li, Y.~Hu, P.~S. Yu, Hgnas++: Efficient architecture search for heterogeneous graph neural networks, IEEE Transactions on Knowledge and Data Engineering 35~(9) (2023) 9448--9461.
\newblock \href {https://doi.org/10.1109/TKDE.2023.3239842} {\path{doi:10.1109/TKDE.2023.3239842}}.

\bibitem{9321762}
K.~C. Tan, L.~Feng, M.~Jiang, Evolutionary transfer optimization - a new frontier in evolutionary computation research, IEEE Computational Intelligence Magazine 16~(1) (2021) 22--33.
\newblock \href {https://doi.org/10.1109/MCI.2020.3039066} {\path{doi:10.1109/MCI.2020.3039066}}.

\bibitem{9756606}
A.~Gupta, L.~Zhou, Y.-S. Ong, Z.~Chen, Y.~Hou, Half a dozen real-world applications of evolutionary multitasking, and more, IEEE Computational Intelligence Magazine 17~(2) (2022) 49--66.
\newblock \href {https://doi.org/10.1109/MCI.2022.3155332} {\path{doi:10.1109/MCI.2022.3155332}}.

\bibitem{9761797}
M.~Shakeri, E.~Miahi, A.~Gupta, Y.-S. Ong, Scalable transfer evolutionary optimization: Coping with big task instances, IEEE Transactions on Cybernetics 53~(10) (2023) 6160--6172.
\newblock \href {https://doi.org/10.1109/TCYB.2022.3164399} {\path{doi:10.1109/TCYB.2022.3164399}}.

\bibitem{qin2022nasbenchgraph}
Y.~Qin, Z.~Zhang, X.~Wang, Z.~Zhang, W.~Zhu, {NAS}-bench-graph: Benchmarking graph neural architecture search, in: Thirty-sixth Conference on Neural Information Processing Systems Datasets and Benchmarks Track, 2022.

\bibitem{pmlr-v70-gilmer17a}
J.~Gilmer, S.~S. Schoenholz, P.~F. Riley, O.~Vinyals, G.~E. Dahl, Neural message passing for quantum chemistry, in: Proceedings of the 34th International Conference on Machine Learning, Vol.~70, PMLR, 2017, pp. 1263--1272.

\bibitem{NIPS2016_04df4d43}
M.~Defferrard, X.~Bresson, P.~Vandergheynst, Convolutional neural networks on graphs with fast localized spectral filtering, in: Advances in Neural Information Processing Systems, Vol.~29, Curran Associates, Inc., 2016.

\bibitem{9336270}
F.~M. Bianchi, D.~Grattarola, L.~Livi, C.~Alippi, Graph neural networks with convolutional arma filters, IEEE Transactions on Pattern Analysis and Machine Intelligence 44~(7) (2022) 3496--3507.
\newblock \href {https://doi.org/10.1109/TPAMI.2021.3054830} {\path{doi:10.1109/TPAMI.2021.3054830}}.

\bibitem{Morris_Ritzert_Fey_Hamilton_Lenssen_Rattan_Grohe_2019}
C.~Morris, M.~Ritzert, M.~Fey, W.~L. Hamilton, J.~E. Lenssen, G.~Rattan, M.~Grohe, Weisfeiler and leman go neural: Higher-order graph neural networks, Proceedings of the AAAI Conference on Artificial Intelligence 33~(01) (2019) 4602--4609.
\newblock \href {https://doi.org/10.1609/aaai.v33i01.33014602} {\path{doi:10.1609/aaai.v33i01.33014602}}.

\bibitem{9378060}
S.~Jiang, P.~Balaprakash, Graph neural network architecture search for molecular property prediction, in: IEEE International Conference on Big Data, IEEE, 2020, pp. 1346--1353.
\newblock \href {https://doi.org/10.1109/BigData50022.2020.9378060} {\path{doi:10.1109/BigData50022.2020.9378060}}.

\bibitem{Li_2019_ICCV}
G.~Li, M.~Muller, A.~Thabet, B.~Ghanem, \href{https://doi.ieeecomputersociety.org/10.1109/ICCV.2019.00936}{{ DeepGCNs: Can GCNs Go As Deep As CNNs? }}, in: 2019 IEEE/CVF International Conference on Computer Vision (ICCV), IEEE Computer Society, Los Alamitos, CA, USA, 2019, pp. 9266--9275.
\newblock \href {https://doi.org/10.1109/ICCV.2019.00936} {\path{doi:10.1109/ICCV.2019.00936}}.
\newline\urlprefix\url{https://doi.ieeecomputersociety.org/10.1109/ICCV.2019.00936}

\bibitem{liu2018darts}
H.~Liu, K.~Simonyan, Y.~Yang, {DARTS}: Differentiable architecture search, in: International Conference on Learning Representations, 2019.

\bibitem{zhao2020probabilistic}
Y.~Zhao, D.~Wang, X.~Gao, R.~Mullins, P.~Lio, M.~Jamnik, Probabilistic dual network architecture search on graphs, arXiv preprint arXiv:2003.09676 (2020).

\bibitem{qin2023multitask}
Y.~Qin, X.~Wang, Z.~Zhang, H.~Chen, W.~Zhu, \href{https://openreview.net/forum?id=TOxpAwp0VE}{Multi-task graph neural architecture search with task-aware collaboration and curriculum}, in: Thirty-seventh Conference on Neural Information Processing Systems, 2023.
\newline\urlprefix\url{https://openreview.net/forum?id=TOxpAwp0VE}

\bibitem{zoph2017neural}
B.~Zoph, Q.~Le, Neural architecture search with reinforcement learning, in: International Conference on Learning Representations, 2017.

\bibitem{zhou2019auto}
K.~Zhou, X.~Huang, Q.~Song, R.~Chen, X.~Hu, Auto-gnn: Neural architecture search of graph neural networks, Frontiers in big Data 5 (2022) 1029307.

\bibitem{ijcai2020p195}
Y.~Gao, H.~Yang, P.~Zhang, C.~Zhou, Y.~Hu, Graph neural architecture search, in: Proceedings of the Twenty-Ninth International Joint Conference on Artificial Intelligence, {IJCAI-20}, International Joint Conferences on Artificial Intelligence Organization, 2020, pp. 1403--1409, main track.
\newblock \href {https://doi.org/10.24963/ijcai.2020/195} {\path{doi:10.24963/ijcai.2020/195}}.

\bibitem{Cai_Wang_Li_Zhang_Zhu_2024}
J.~Cai, X.~Wang, H.~Li, Z.~Zhang, W.~Zhu, \href{https://ojs.aaai.org/index.php/AAAI/article/view/28663}{Multimodal graph neural architecture search under distribution shifts}, Proceedings of the AAAI Conference on Artificial Intelligence 38~(8) (2024) 8227--8235.
\newblock \href {https://doi.org/10.1609/aaai.v38i8.28663} {\path{doi:10.1609/aaai.v38i8.28663}}.
\newline\urlprefix\url{https://ojs.aaai.org/index.php/AAAI/article/view/28663}

\bibitem{784219}
X.~Yao, Evolving artificial neural networks, Proceedings of the IEEE 87~(9) (1999) 1423--1447.
\newblock \href {https://doi.org/10.1109/5.784219} {\path{doi:10.1109/5.784219}}.

\bibitem{9508774}
Y.~Liu, Y.~Sun, B.~Xue, M.~Zhang, G.~G. Yen, K.~C. Tan, A survey on evolutionary neural architecture search, IEEE Transactions on Neural Networks and Learning Systems 34~(2) (2023) 550--570.
\newblock \href {https://doi.org/10.1109/TNNLS.2021.3100554} {\path{doi:10.1109/TNNLS.2021.3100554}}.

\bibitem{10004638}
Z.~Lu, R.~Cheng, Y.~Jin, K.~C. Tan, K.~Deb, Neural architecture search as multiobjective optimization benchmarks: Problem formulation and performance assessment, IEEE Transactions on Evolutionary Computation 28~(2) (2024) 323--337.
\newblock \href {https://doi.org/10.1109/TEVC.2022.3233364} {\path{doi:10.1109/TEVC.2022.3233364}}.

\bibitem{10.1145/3449639.3459318}
M.~Nunes, P.~M. Fraga, G.~L. Pappa, Fitness landscape analysis of graph neural network architecture search spaces, in: Proceedings of the Genetic and Evolutionary Computation Conference, GECCO '21, ACM, 2021, p. 876–884.
\newblock \href {https://doi.org/10.1145/3449639.3459318} {\path{doi:10.1145/3449639.3459318}}.

\bibitem{LIU2023110485}
Y.~Liu, J.~Liu, \href{https://www.sciencedirect.com/science/article/pii/S1568494623005033}{A surrogate evolutionary neural architecture search algorithm for graph neural networks}, Applied Soft Computing 144 (2023) 110485.
\newblock \href {https://doi.org/https://doi.org/10.1016/j.asoc.2023.110485} {\path{doi:https://doi.org/10.1016/j.asoc.2023.110485}}.
\newline\urlprefix\url{https://www.sciencedirect.com/science/article/pii/S1568494623005033}

\bibitem{10681642}
C.~Wang, J.~Zhao, L.~Li, L.~Jiao, F.~Liu, S.~Yang, Automatic graph topology-aware transformer, IEEE Transactions on Neural Networks and Learning Systems (2024) 1--15\href {https://doi.org/10.1109/TNNLS.2024.3440269} {\path{doi:10.1109/TNNLS.2024.3440269}}.

\bibitem{xue2022}
X.~Xue, C.~Yang, L.~Feng, K.~Zhang, L.~Song, K.~C. Tan, {How to Exploit Optimization Experience? Revisiting Evolutionary Sequential Transfer Optimization: Part A - Benchmark Problems}, techrxiv preprint techrxiv.21694754.v1 (2022).
\newblock \href {https://doi.org/10.36227/techrxiv.21694754.v1} {\path{doi:10.36227/techrxiv.21694754.v1}}.

\bibitem{7161358}
A.~Gupta, Y.-S. Ong, L.~Feng, Multifactorial evolution: Toward evolutionary multitasking, IEEE Transactions on Evolutionary Computation 20~(3) (2016) 343--357.
\newblock \href {https://doi.org/10.1109/TEVC.2015.2458037} {\path{doi:10.1109/TEVC.2015.2458037}}.

\bibitem{MA2023110027}
X.~Ma, M.~Xu, Y.~Yu, H.~Liu, Y.~Wang, L.~Wang, Y.~Qi, J.~Xiong, \href{https://www.sciencedirect.com/science/article/pii/S0950705122011200}{Enhancing evolutionary multitasking optimization by leveraging inter-task knowledge transfers and improved evolutionary operators}, Knowledge-Based Systems 259 (2023) 110027.
\newblock \href {https://doi.org/https://doi.org/10.1016/j.knosys.2022.110027} {\path{doi:https://doi.org/10.1016/j.knosys.2022.110027}}.
\newline\urlprefix\url{https://www.sciencedirect.com/science/article/pii/S0950705122011200}

\bibitem{GAO2024111530}
F.~Gao, W.~Gao, L.~Huang, S.~Zhang, M.~Gong, L.~Wang, \href{https://www.sciencedirect.com/science/article/pii/S0950705124001655}{Effective transferred knowledge identified by bipartite graph for multiobjective multitasking optimization}, Knowledge-Based Systems 290 (2024) 111530.
\newblock \href {https://doi.org/https://doi.org/10.1016/j.knosys.2024.111530} {\path{doi:https://doi.org/10.1016/j.knosys.2024.111530}}.
\newline\urlprefix\url{https://www.sciencedirect.com/science/article/pii/S0950705124001655}

\bibitem{10026148}
C.~Wang, J.~Zhao, L.~Li, L.~Jiao, J.~Liu, K.~Wu, A multi-transformation evolutionary framework for influence maximization in social networks, IEEE Computational Intelligence Magazine 18~(1) (2023) 52--67.
\newblock \href {https://doi.org/10.1109/MCI.2022.3222050} {\path{doi:10.1109/MCI.2022.3222050}}.

\bibitem{xue20221}
X.~Xue, C.~Yang, L.~Feng, K.~Zhang, L.~Song, K.~C. Tan, {How to Exploit Optimization Experience? Revisiting Evolutionary Sequential Transfer Optimization: Part B - Algorithm Analysis}, techrxiv preprint techrxiv.21694832.v1 (2022).
\newblock \href {https://doi.org/10.36227/techrxiv.21694832.v1} {\path{doi:10.36227/techrxiv.21694832.v1}}.

\bibitem{9385398}
C.~Wang, J.~Liu, K.~Wu, Z.~Wu, Solving multitask optimization problems with adaptive knowledge transfer via anomaly detection, IEEE Transactions on Evolutionary Computation 26~(2) (2022) 304--318.
\newblock \href {https://doi.org/10.1109/TEVC.2021.3068157} {\path{doi:10.1109/TEVC.2021.3068157}}.

\bibitem{9756594}
C.~Wang, K.~Wu, J.~Liu, Evolutionary multitasking auc optimization [research frontier], IEEE Computational Intelligence Magazine 17~(2) (2022) 67--82.
\newblock \href {https://doi.org/10.1109/MCI.2022.3155325} {\path{doi:10.1109/MCI.2022.3155325}}.

\bibitem{WANG2021107190}
X.~Wang, Y.~Jin, S.~Schmitt, M.~Olhofer, R.~Allmendinger, \href{https://www.sciencedirect.com/science/article/pii/S0950705121004524}{Transfer learning based surrogate assisted evolutionary bi-objective optimization for objectives with different evaluation times}, Knowledge-Based Systems 227 (2021) 107190.
\newblock \href {https://doi.org/https://doi.org/10.1016/j.knosys.2021.107190} {\path{doi:https://doi.org/10.1016/j.knosys.2021.107190}}.
\newline\urlprefix\url{https://www.sciencedirect.com/science/article/pii/S0950705121004524}

\bibitem{10.1162/evco_a_00300}
X.~Wang, Y.~Jin, S.~Schmitt, M.~Olhofer, Transfer learning based co-surrogate assisted evolutionary bi-objective optimization for objectives with non-uniform evaluation times, Evolutionary Computation 30~(2) (2022) 221--251.
\newblock \href {https://doi.org/10.1162/evco_a_00300} {\path{doi:10.1162/evco_a_00300}}.

\bibitem{9644585}
X.~Xue, C.~Yang, Y.~Hu, K.~Zhang, Y.-M. Cheung, L.~Song, K.~C. Tan, Evolutionary sequential transfer optimization for objective-heterogeneous problems, IEEE Transactions on Evolutionary Computation 26~(6) (2022) 1424--1438.
\newblock \href {https://doi.org/10.1109/TEVC.2021.3133874} {\path{doi:10.1109/TEVC.2021.3133874}}.

\bibitem{Nomura_Watanabe_Akimoto_Ozaki_Onishi_2021}
M.~Nomura, S.~Watanabe, Y.~Akimoto, Y.~Ozaki, M.~Onishi, Warm starting cma-es for hyperparameter optimization, Proceedings of the AAAI Conference on Artificial Intelligence 35~(10) (2021) 9188--9196.
\newblock \href {https://doi.org/10.1609/aaai.v35i10.17109} {\path{doi:10.1609/aaai.v35i10.17109}}.

\bibitem{9950429}
N.~Zhang, A.~Gupta, Z.~Chen, Y.-S. Ong, Multitask neuroevolution for reinforcement learning with long and short episodes, IEEE Transactions on Cognitive and Developmental Systems (2022) 1--1\href {https://doi.org/10.1109/TCDS.2022.3221805} {\path{doi:10.1109/TCDS.2022.3221805}}.

\bibitem{7879282}
L.~Feng, Y.-S. Ong, S.~Jiang, A.~Gupta, Autoencoding evolutionary search with learning across heterogeneous problems, IEEE Transactions on Evolutionary Computation 21~(5) (2017) 760--772.
\newblock \href {https://doi.org/10.1109/TEVC.2017.2682274} {\path{doi:10.1109/TEVC.2017.2682274}}.

\bibitem{10342789}
X.~Xue, C.~Yang, L.~Feng, K.~Zhang, L.~Song, K.~C. Tan, Solution transfer in evolutionary optimization: An empirical study on sequential transfer, IEEE Transactions on Evolutionary Computation (2023) 1--1\href {https://doi.org/10.1109/TEVC.2023.3339506} {\path{doi:10.1109/TEVC.2023.3339506}}.

\bibitem{10.1145/3594805.3607137}
E.~O. Scott, K.~A. De~Jong, \href{https://doi.org/10.1145/3594805.3607137}{First complexity results for evolutionary knowledge transfer}, in: Proceedings of the 17th ACM/SIGEVO Conference on Foundations of Genetic Algorithms, FOGA '23, Association for Computing Machinery, New York, NY, USA, 2023, p. 140–151.
\newblock \href {https://doi.org/10.1145/3594805.3607137} {\path{doi:10.1145/3594805.3607137}}.
\newline\urlprefix\url{https://doi.org/10.1145/3594805.3607137}

\bibitem{lee2021rapid}
H.~Lee, E.~Hyung, S.~J. Hwang, Rapid neural architecture search by learning to generate graphs from datasets, in: International Conference on Learning Representations, 2021.

\bibitem{9328602}
Z.~Lu, G.~Sreekumar, E.~Goodman, W.~Banzhaf, K.~Deb, V.~N. Boddeti, Neural architecture transfer, IEEE Transactions on Pattern Analysis and Machine Intelligence 43~(9) (2021) 2971--2989.
\newblock \href {https://doi.org/10.1109/TPAMI.2021.3052758} {\path{doi:10.1109/TPAMI.2021.3052758}}.

\bibitem{wang2022automated}
X.~Wang, Z.~Zhang, W.~Zhu, Automated graph machine learning: Approaches, libraries and directions, arXiv preprint arXiv:2201.01288 (2022).

\bibitem{pmlr-v97-astudillo19a}
R.~Astudillo, P.~Frazier, {B}ayesian optimization of composite functions, in: Proceedings of the 36th International Conference on Machine Learning, Vol.~97, PMLR, 2019, pp. 354--363.

\bibitem{8836854}
J.~Luo, J.~Feng, R.~Jin, A new approach to building the gaussian process model for expensive multi-objective optimization, in: 2019 9th International Conference on Information Science and Technology, IEEE, 2019, pp. 374--379.
\newblock \href {https://doi.org/10.1109/ICIST.2019.8836854} {\path{doi:10.1109/ICIST.2019.8836854}}.

\bibitem{LIU2018102}
H.~Liu, J.~Cai, Y.-S. Ong, Remarks on multi-output gaussian process regression, Knowledge-Based Systems 144 (2018) 102--121.

\bibitem{NEURIPS2021_a0d3973a}
W.~J. Maddox, M.~Balandat, A.~G. Wilson, E.~Bakshy, Bayesian optimization with high-dimensional outputs, in: Advances in Neural Information Processing Systems, Vol.~34, Curran Associates, Inc., 2021.

\bibitem{jakkala2021deep}
K.~Jakkala, Deep gaussian processes: A survey, arXiv preprint arXiv:2106.12135 (2021).

\bibitem{996017}
K.~Deb, A.~Pratap, S.~Agarwal, T.~Meyarivan, A fast and elitist multiobjective genetic algorithm: Nsga-ii, IEEE Transactions on Evolutionary Computation 6~(2) (2002) 182--197.
\newblock \href {https://doi.org/10.1109/4235.996017} {\path{doi:10.1109/4235.996017}}.

\bibitem{kipf2016variational}
T.~N. Kipf, M.~Welling, Variational graph auto-encoders, arXiv preprint arXiv:1611.07308 (2016).

\bibitem{NEURIPS2019_e205ee2a}
M.~Zhang, S.~Jiang, Z.~Cui, R.~Garnett, Y.~Chen, D-vae: A variational autoencoder for directed acyclic graphs, in: Advances in Neural Information Processing Systems, Vol.~32, Curran Associates, Inc., 2019.

\bibitem{pmlr-v198-zhu22a}
Y.~Zhu, Y.~Du, Y.~Wang, Y.~Xu, J.~Zhang, Q.~Liu, S.~Wu, \href{https://proceedings.mlr.press/v198/zhu22a.html}{A survey on deep graph generation: Methods and applications}, in: B.~Rieck, R.~Pascanu (Eds.), Proceedings of the First Learning on Graphs Conference, Vol. 198 of Proceedings of Machine Learning Research, PMLR, 2022, pp. 47:1--47:21.
\newline\urlprefix\url{https://proceedings.mlr.press/v198/zhu22a.html}

\bibitem{pmlr-v115-li20c}
L.~Li, A.~Talwalkar, Random search and reproducibility for neural architecture search, in: Proceedings of The 35th Uncertainty in Artificial Intelligence Conference, Vol. 115, PMLR, 2020, pp. 367--377.

\bibitem{Real_Aggarwal_Huang_Le_2019}
E.~Real, A.~Aggarwal, Y.~Huang, Q.~V. Le, Regularized evolution for image classifier architecture search, Proceedings of the AAAI Conference on Artificial Intelligence 33~(01) (2019) 4780--4789.

\bibitem{pmlr-v80-xu18c}
K.~Xu, C.~Li, Y.~Tian, T.~Sonobe, K.-i. Kawarabayashi, S.~Jegelka, Representation learning on graphs with jumping knowledge networks, in: Proceedings of the 35th International Conference on Machine Learning, Vol.~80, PMLR, 2018, pp. 5453--5462.

\bibitem{Pei2020Geom}
H.~Pei, B.~Wei, K.~C.-C. Chang, Y.~Lei, B.~Yang, Geom-gcn: Geometric graph convolutional networks, in: International Conference on Learning Representations, 2020.

\bibitem{Zeng2020GraphSAINT}
H.~Zeng, H.~Zhou, A.~Srivastava, R.~Kannan, V.~Prasanna, Graphsaint: Graph sampling based inductive learning method, in: International Conference on Learning Representations, 2020.

\bibitem{zhao2020simplifying}
H.~Zhao, L.~Wei, Q.~Yao, Simplifying architecture search for graph neural network, arXiv preprint arXiv:2008.11652 (2020).

\bibitem{1688440}
C.~Fonseca, L.~Paquete, M.~Lopez-Ibanez, An improved dimension-sweep algorithm for the hypervolume indicator, in: 2006 IEEE International Conference on Evolutionary Computation, IEEE, 2006, pp. 1157--1163.
\newblock \href {https://doi.org/10.1109/CEC.2006.1688440} {\path{doi:10.1109/CEC.2006.1688440}}.

\end{thebibliography}






\end{document}